\RequirePackage{fix-cm}

\documentclass{alaya}
\usepackage[sc]{mathpazo}
\usepackage[numbers]{natbib}
\usepackage{environ}

\usepackage{amsmath,amsfonts,bm}

\def\eqref#1{(\ref{#1})}

\def\1{\bm{1}}

\DeclareMathAlphabet{\mathsfit}{\encodingdefault}{\sfdefault}{m}{sl}
\SetMathAlphabet{\mathsfit}{bold}{\encodingdefault}{\sfdefault}{bx}{n}

\newcommand{\gtS}[1]{S_{#1}}                                   %
\newcommand{\predS}[1]{\hat S_{#1}}                            %
\newcommand{\eparts}[1]{\mathcal{M}_{#1}}                      %
\newcommand{\restr}[2]{#1\!\left[#2\right]}                    %
\newcommand{\grid}{\mathcal{G}}                                %
\newcommand{\imgdom}{\Omega}                                   %
\newcommand{\vox}[1]{\mathcal{V}\!\left(#1\right)}             %
\newcommand{\dil}[1]{\mathcal{D}\!\left(#1\right)}             %
\newcommand{\editR}[1]{E_{#1}}                                 %
\newcommand{\keepR}[1]{U_{#1}}                                 %
\newcommand{\view}{v}                                          %
\newcommand{\proj}[2]{\Pi^{\view}\!\left(#1;#2\right)}         %
\newcommand{\editRimg}[1]{E^{\view}_{#1}}                      %
\newcommand{\keepRimg}[1]{U^{\view}_{#1}}                      %
\newcommand{\pts}[1]{\mathcal{Q}\!\left(#1\right)}             %
\newcommand{\keepRcum}[1]{U^{0}_{#1}}                          %
\newcommand{\IoUon}[3]{\mathrm{IoU}_{#1}\!\left(#2,#3\right)}
\newcommand{\CDon}[3]{\mathrm{CD}_{#1}\!\left(#2,#3\right)}
\newcommand{\Fon}[3]{\mathrm{F}^{\tau}_{#1}\!\left(#2,#3\right)}
\newcommand{\PSNRon}[3]{\mathrm{PSNR}_{#1}\!\left(#2,#3\right)}
\newcommand{\SSIMon}[3]{\mathrm{SSIM}_{#1}\!\left(#2,#3\right)}
\newcommand{\LPIPSon}[3]{\mathrm{LPIPS}_{#1}\!\left(#2,#3\right)}
\newcommand{\DINOon}[3]{\mathrm{DINO}_{#1}\!\left(#2,#3\right)}
\newcommand{\MAEon}[3]{\mathrm{MAE}_{#1}\!\left(#2,#3\right)}
\newcommand{\avg}[1]{\operatorname*{mean}_{#1}}
\newcommand{\Recallon}[3]{\mathrm{Recall}_{#1}\!\left(#2,#3\right)}   %

\hypersetup{breaklinks}
\usepackage{url}
\usepackage{graphicx}
\usepackage{booktabs}
\usepackage{tabularx}
\usepackage[export]{adjustbox}
\usepackage{rotating}
\usepackage{enumitem}
\usepackage{array}
\usepackage{xcolor}
\usepackage{amsmath,amssymb}
\usepackage{tikz,pgfplots}
\pgfplotsset{compat=1.18,table/search path={.}}
\usepgfplotslibrary{groupplots,fillbetween}
\usetikzlibrary{calc,arrows.meta,shadows.blur}
\definecolor{c1}{HTML}{7DAEE0}\definecolor{c2}{HTML}{E9C46A}\definecolor{c3}{HTML}{299D8F}\definecolor{c4}{HTML}{D87659}\definecolor{c5}{HTML}{8A78AD}
\definecolor{c-code}{HTML}{2B2B2B}\definecolor{pl-ink}{HTML}{2B2B2B}\definecolor{pl-mute}{HTML}{7A7A7A}\definecolor{pl-line}{HTML}{D9D6D0}
\definecolor{op-add}{HTML}{7DAEE0}\definecolor{op-rem}{HTML}{299D8F}\definecolor{op-rep}{HTML}{E9C46A}\definecolor{op-tex}{HTML}{D87659}
\pgfplotsset{eh-bar/.style={
 font=\sffamily\fontsize{6.4}{7.4}\selectfont,
 title style={font=\sffamily\bfseries\fontsize{7}{8}\selectfont,text=pl-ink,yshift=-4pt},
 label style={font=\sffamily\fontsize{6.4}{7.4}\selectfont,text=pl-ink},
 tick label style={font=\sffamily\fontsize{6.4}{7.4}\selectfont,text=pl-ink},
 axis line style={draw=pl-ink,line width=.6pt},
 axis x line*=bottom,axis y line*=left,
 tick style={draw=pl-ink,line width=.6pt},tick align=outside,major tick length=2.5pt,grid=none,
 legend style={font=\sffamily\fontsize{6.4}{7.4}\selectfont,draw=none,fill=none,
   inner sep=0pt,/tikz/every even column/.append style={column sep=10pt}},
 legend cell align=left,
 legend image code/.code={\draw[##1,draw=none] (0cm,-.04cm) rectangle (.18cm,.08cm);},
 clip=false
}}
\definecolor{eh-main}{HTML}{B395BD}%

\definecolor{pastelA}{HTML}{F4C3BD}\definecolor{pastelB}{HTML}{F7D3AC}\definecolor{pastelC}{HTML}{F4E5A8}
\definecolor{pastelD}{HTML}{DDEAA9}\definecolor{pastelE}{HTML}{B9E4BF}\definecolor{pastelF}{HTML}{ADE1DC}
\definecolor{pastelG}{HTML}{B3D5F0}\definecolor{pastelH}{HTML}{C2C5EC}\definecolor{pastelI}{HTML}{CFBCE9}
\definecolor{pastelJ}{HTML}{E2BCDF}
\definecolor{chainbar}{HTML}{ECECF1}

\newsavebox{\ehbox}
\newcommand{\ehalaya}[3][0.85]{\sbox\ehbox{#3}\adjustbox{scale=#2,max width=\linewidth,max totalheight=#1\textheight}{\usebox\ehbox}}
\newcommand{\ehfit}[2][1.25]{\ehalaya{#1}{#2}}
\newcommand{\ehfitplain}[2][1.25]{\ehalaya{#1}{#2}}
\newcommand{\ehfitpage}[2][1.25]{\ehalaya[0.925]{#1}{#2}}
\def\ehtabsepval{}
\AtBeginEnvironment{tabular}{\ifx\ehtabsepval\empty\else\setlength{\tabcolsep}{\ehtabsepval}\fi}
\newcommand{\ehtextbf}[1]{{\bfseries #1}}
\AtBeginEnvironment{tabular}{\let\textbf\ehtextbf}\AtBeginEnvironment{tabularx}{\let\textbf\ehtextbf}
\newcommand{\ehsep}[1]{\def\ehtabsepval{#1}}
\newcommand{\nChains}{457}
\newcommand{\nHosts}{252}
\newcommand{\nEdits}{2755}
\newcommand{\maxTurns}{30}
\newcommand{\minTurns}{3}
\newcommand{\nMaterialChains}{132}
\newcommand{\cohortChains}{416}
\newcommand{\cohortHosts}{228}
\newcommand{\gtResetChains}{200}
\newcommand{\matrepOwnOverGT}{0.31}
\newcommand{\nEvalChains}{457}
\newcommand{\nEvalEdits}{2755}
\newcommand{\turnFigChains}{135}\newcommand{\turnFigHosts}{103}%
\newcommand{\gtResetFixedChains}{147}

\newcommand{\secPartflow}{26 (32)}
\newcommand{\secNano}{22 (29)}
\newcommand{\secEditformer}{20 (26)}
\newcommand{\secVox}{49 (80)}

\newcommand{\llmChains}{55}
\newcommand{\llmTurns}{416}
\newcommand{\llmOpusIF}{0.62}
\newcommand{\llmOpusCCprev}{0.98}
\newcommand{\llmOpusLPIPSW}{0.065}
\newcommand{\llmOpusIFadd}{0.49}
\newcommand{\llmOpusIFrem}{0.99}
\newcommand{\llmOpusIFrep}{0.52}
\newcommand{\llmNoopLPIPSW}{0.154}
\newcommand{\llmGenIFmax}{0.36}
\newcommand{\llmGenCCmax}{0.90}
\newcommand{\llmGLMCCprev}{0.95}
\newcommand{\llmGLMLPIPSW}{0.136}
\newcommand{\llmGLMIFadd}{0.10}
\newcommand{\llmGLMIFrem}{0.88}
\newcommand{\llmGLMIFrep}{0.17}

\newcommand{\amortSolSec}{87}
\newcommand{\amortSolPninety}{183}
\newcommand{\amortSolTokIn}{848k}
\newcommand{\amortSolTokOut}{2.67k}
\newcommand{\amortSolCalls}{7.8}
\newcommand{\amortAstraSec}{95}
\newcommand{\amortAstraPninety}{143}
\newcommand{\amortAstraTokIn}{325k}
\newcommand{\amortAstraTokOut}{2.36k}
\newcommand{\amortAstraCalls}{5.8}
\newcommand{\amortOpusSec}{149}
\newcommand{\amortOpusPninety}{291}
\newcommand{\amortOpusTokIn}{1216k}
\newcommand{\amortOpusTokOut}{9.20k}
\newcommand{\amortOpusCalls}{10.9}
\newcommand{\amortFableSec}{258}
\newcommand{\amortFablePninety}{557}
\newcommand{\amortFableTokIn}{1555k}
\newcommand{\amortFableTokOut}{18.26k}
\newcommand{\amortFableCalls}{9.3}
\newcommand{\amortGLMSec}{358}
\newcommand{\amortGLMPninety}{619}
\newcommand{\amortGLMTokIn}{680k}
\newcommand{\amortGLMTokOut}{4.66k}
\newcommand{\amortGLMCalls}{12.8}
\newcommand{\amortDeepSeekSec}{358}
\newcommand{\amortDeepSeekPninety}{1036}
\newcommand{\amortDeepSeekTokIn}{1697k}
\newcommand{\amortDeepSeekTokOut}{14.13k}
\newcommand{\amortDeepSeekCalls}{15.9}

\title{EditHero: A Benchmark for Long-Horizon Part-Level 3D Editing and Vibe Modeling}

\author[1,2,*]{Ruihan Yu}
\author[3,4,*]{Yu-Ju Tsai}
\author[1,2]{Muyao Niu}
\author[1,2]{Runyi Li}
\author[1,2]{Lian Fu}
\author[2]{Hanqing Liu}
\author[1]{Zheng-Hui Huang}
\author[1,2]{Yonghao Yu}
\author[1,2]{Sho Kuno}
\author[4,\dagger]{Ming-Hsuan Yang}
\author[1,\dagger]{Kaipeng Zhang}
\author[1,\dagger]{Zhixiang Wang}
\affiliation[1]{Alaya Lab}
\affiliation[2]{The University of Tokyo}
\affiliation[3]{Institute of Science Tokyo}
\affiliation[4]{University of California, Merced}
\contribution[*]{Equal contribution}
\contribution[\dagger]{Corresponding authors}
\project{\url{https://alaya-lab.github.io/EditHero/}}
\code{\url{https://github.com/AlayaLab/EditHero}}
\correspondence{Zhixiang Wang, Kaipeng Zhang, Ming-Hsuan Yang}
\date{\today}
\let\alayaabstract\abstract\let\abstract\relax
\NewEnviron{abstract}{\global\let\alayaabstractbody\BODY}
\begin{abstract}
3D editing methods are usually tested on a single edit, yet an asset is built through a long sequence of revisions, each of which must implement the requested change while leaving everything else unchanged. We introduce EditHero, to our knowledge the first benchmark for long-horizon, part-level 3D editing, with natural-language instructions and target images for both geometry and texture. A deterministic assembly engine produces the exact target after every edit, and every sequence is reviewed by hand. We use EditHero to compare 2 opposite approaches to 3D editing. Non-agentic methods operate top down, regenerating the object from a learned 3D representation and inferring what to keep. In contrast, LLM/VLM agents operate bottom up, editing through code that inspects the mesh and rewrites only the parts required by instructions. The non-agentic methods often miss the requested change and disturb regions that should stay fixed. Most LLMs follow instructions more closely, and all of them preserve the unedited parts better, but each of their edits takes minutes. We will release the engine and the edit sequences to support research on reliable iterative 3D editing.
\end{abstract}

\expandafter\alayaabstract\expandafter{\alayaabstractbody}

\begin{document}
\maketitle

\input{sec/fig_teaser}

\section{Introduction}
\label{sec:intro}

A 3D artist rarely makes one edit and stops. Assets in art and game production evolve through successive editing and revisions: each turn of modification changes part of the current asset while preserving earlier work. The same expectation underlies the recent concept of ``vibe modeling'', in which users shape assets through an iterative natural-language conversation~\citep{embodiedgen2,ye2025shapellmomni,llmr}. Yet text- and image-guided 3D editing methods~\citep{voxhammer,nano3d,editformer,partflow} are typically evaluated on a single edit from a clean source. It remains unclear whether they can edit their own outputs repeatedly without losing earlier changes or degrading the asset.

Evaluating this ability requires a reference after every turn. Existing paired datasets~\citep{editformer,nano3d,partflow} only provide isolated edits. Their targets come from either learned editing models or changes to poses and semantic parts. Neither design can tell us whether a method preserves earlier changes across a sequence of instructions.

\begin{figure}[t]
\centering
\ehfitplain{\input{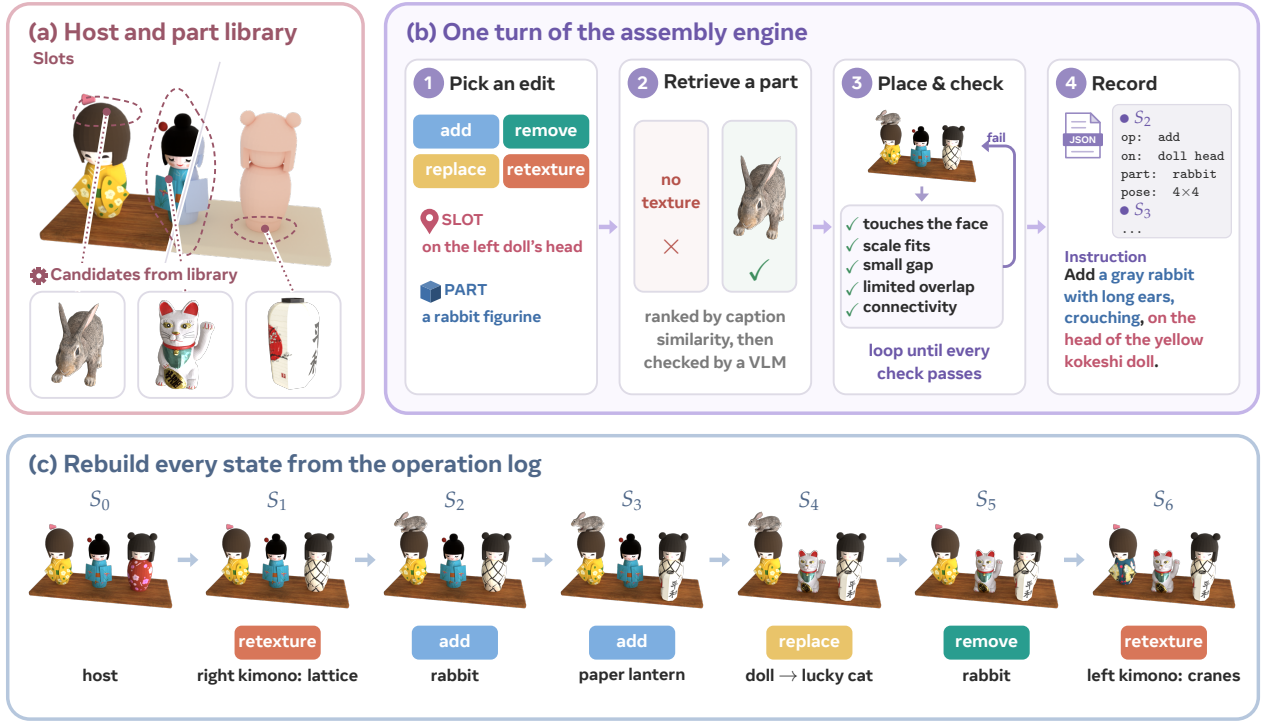}}
\caption{\textbf{Constructing EditHero.} (a) A host object is divided into named slots, and candidate parts are drawn from a captioned library. (b) The addition shown illustrates edit selection, part retrieval, visual screening and placement checks. Each accepted edit is recorded in the chain log, and reviewers refine the drafted instructions. (c) With the part library, the log rebuilds the exact target of any turn. A method under test never sees the log. It receives the instruction and the target render of each turn. The example combines recorded turns from 4 chains on the same host; all images are renders. Colors indicate the 4 edit operations.}
\label{fig:pipeline}
\end{figure}

We introduce EditHero, a benchmark of \emph{part-level edit chains with exact ground truth} (Figure~\ref{fig:pipeline}). The benchmark is built with the help of a data engine, which assembles library parts onto segmented host objects with 4 operations (\textbf{add}, \textbf{remove}, \textbf{replace} and \textbf{retexture}) and records every operation in a JSON file. Each entry is a single atomic operation, so replaying the file with the part library rebuilds the exact state after any turn. After human review, EditHero contains \nChains{} chains and \nEdits{} edits across \nHosts{} objects, with up to \maxTurns{} turns per chain (Section~\ref{sec:benchmark}).

We test 2 families of methods on this long-horizon 3D editing task, and they approach an edit in opposite ways. Non-agentic methods work top down. They regenerate the edited object from a learned 3D representation, such as the latents of a diffusion model, so every turn changes every corner of the asset, more or less. In contrast, LLM/VLM agents work bottom up. An LLM or VLM inspects the current mesh and writes code that changes only the named part. This keeps everything else untouched by construction and forces every new part to be built from primitives. We compare 4 non-agentic methods~\citep{partflow,nano3d,editformer,voxhammer} with 6 LLM/VLM agents under a \emph{self-rollout} protocol (Section~\ref{sec:protocol}), in which every edit starts from the method's own previous output. We also design 2 metrics to measure the differences between these methods, including \textbf{instruction following (IF)} inside the edit region and \textbf{content consistency (CC)} outside it.

The results indicate that agentic methods have superior editing ability. The non-agentic methods often miss the requested change as early as the first turn, and errors accumulate along a chain. The colors of their results also shift away from the target. The agents behave differently. Most agents follow instructions more closely than non-agentic methods, and all agents keep the untouched parts better. However, the agents' speed is not so satisfying, as a single edit takes minutes rather than seconds (Sections~\ref{sec:experiments} and~\ref{sec:agentic}).
Our contributions are:
\begin{itemize}[leftmargin=*,itemsep=1pt,topsep=1pt]
\item \textbf{Generation engine.} 
We construct a data synthesis engine that produces part-level edit chains with an exact target after every turn.
\item \textbf{Benchmark.}
We propose \textit{EditHero}, a human-reviewed benchmark for long-horizon part-level 3D editing.
\item \textbf{Evaluation framework.}
We evaluate 4 non-agentic methods and 6 LLM/VLM agents under self-rollout, with region-specific measures of instruction following and preservation.
\end{itemize}
\section{Related Work}
\label{sec:related}

\noindent \textbf{3D editing.}
3D editing has followed the available generative models. Early methods optimized a NeRF or a Gaussian splat under a 2D diffusion prior~\citep{haque2023instructnerf2nerf,chen2024gaussianeditor,chen2024dge}, and later ones edit rendered views or an image and then reconstruct the object~\citep{barda2025instant3dit,hsiao2026vecsetedit}. Part-level and sketch-based methods let a user refine an object step by step~\citep{dong2024interactive3d,sella2026proxe,sketch2cad,meshpad}. Native 3D generators such as TRELLIS~\citep{trellis} made it possible to edit a 3D latent directly. Methods trained on paired edits feed the source latent to the generator through an added branch, much as ControlNet~\citep{controlnet} conditions an image model. PartFlow~\citep{partflow} adds a source-control branch, and 3DEditFormer~\citep{editformer} uses dual-guidance attention. Training-free methods reuse a pretrained generator instead. VoxHammer~\citep{voxhammer} inverts the source and reuses its cached latents and attention features in the preserved region, and Nano3D~\citep{nano3d} follows the difference between the source-conditioned and target-conditioned flows, using the FlowEdit formulation~\citep{kulikov2025flowedit}. We evaluate PartFlow, 3DEditFormer, VoxHammer and Nano3D as representatives. Each was proposed and evaluated for one edit at a time, and each uses generative latents to synthesize edited content. EditHero instead asks them to keep editing their own output, as an artist would, and measures what regeneration does to the parts not required by any instruction. Appendix~\ref{app:related} describes these methods in more detail and lists the others.

\noindent \textbf{3D editing benchmarks.}
Most text- and image-guided 3D editing benchmarks score one instruction at a time. Some pair a source asset with a target asset, obtained through image-guided generation~\citep{editformer,nano3d,ma2025steer3d} or pose and part changes~\citep{editformer,partflow,preintner2026threedmorph}. Others have no target asset and score region preservation and alignment to an edited image instead~\citep{voxhammer,zhou2026anchorflow,gat2026shapeup,nie2026editflow3d}, and scene-level benchmarks follow the same pattern~\citep{patro2026editbench3d,liu2025edit3r,zhu2026jointedit3d}. The closest multi-step settings are agent benchmarks such as BlenderGym~\citep{gu2025blendergym} and 3DCodeBench~\citep{gao2026threedcodebench}, which let an agent iterate toward one fixed goal. EZBlender~\citep{wang2026ezblender} instead bundles several edits into one prompt. To our knowledge, EditHero is the first benchmark to evaluate different natural-language part edits in sequence on one asset, with an exact 3D reference after every turn (Appendix Table~\ref{tab:bench-compare}).

\noindent \textbf{LLM/VLM agents for 3D generation and editing.}
LLM agents can modify 3D assets through code. They write programs or call tools to build scenes~\citep{threedgpt,holodeck} and edit Blender scenes from renders~\citep{blenderalchemy,lu2025ll3m}. Related systems generate objects and CAD programs~\citep{meshcoder,text2cad,cadrecode}. Recent systems treat the program itself as the asset and revise it under visual feedback~\citep{procedura,vibeworlding,embodiedgen2}, aiming to keep edits local and preserve the rest of the state. Our data engine also records edits as operations that can be replayed to reconstruct each state. We evaluate such agents on the mesh, turn by turn and with the same metrics as the non-agentic methods (Section~\ref{sec:agentic}).

\section{Benchmark Construction}
\label{sec:benchmark}

\noindent\textbf{Task.}
We hope to do multi-turn editing starting from a textured, part-segmented \emph{host} object $S_0$ with text or image conditions. We define 4 operations, \textbf{add}, \textbf{remove}, \textbf{replace} and \textbf{retexture}. Each turn specifies one operation on a part through a text instruction and a fixed-view render of the target state. An \textbf{add} attaches a part to a free contact face, and a \textbf{remove} deletes a named part, which may consist of several pieces. A \textbf{replace} swaps a named part for another at the same location, while a \textbf{retexture} changes how a part looks without touching its geometry. All other parts should remain unchanged. After each turn, the method returns a 3D object that becomes its input at the next turn. Text instructions may refer to earlier edits. For example, a part added at one turn can be removed a few turns later, so a method may want to keep track of what it has already built.

\noindent\textbf{Exact ground truth by construction.}
Evaluating a chain requires a reliable target at every turn. Some existing paired datasets obtain their targets from image-guided generative pipelines~\citep{editformer,nano3d}, and such a target can itself contain changes nobody asked for. We instead assemble each target from parts, much as CLEVR and Kubric generate their scenes programmatically~\citep{clevr,kubric}. For an addition or replacement, the engine retrieves caption-matched parts from a library built from PartVerse-XL~\citep{fullpart}, Objaverse-XL~\citep{objaversexl} and HY3D-Bench~\citep{hy3dbench}. It then screens the candidates with a vision-language model (VLM) and tests their placement for contact, scale, connectivity, and interpenetration. Removal deletes the part occupying a named slot. For a retexture, the engine renders the slot and restyles the render with an image editing model (Qwen-Image-Edit~\citep{qwenimage}). The texturing module of TRELLIS.2~\citep{trellis2} then textures the unchanged slot geometry from the restyled image. Each chain is stored as a JSON file that logs every accepted operation with its library part, pose and texture. Together with the part library, the log rebuilds any state $S_k$ exactly and records what changed at each turn.

\noindent\textbf{Human review.}
Reviewers inspect renders of every turn for incorrect parts, orientation, placement, scale, ambiguous instructions, and broken dependencies. They correct flagged cases by editing the logged operations and rebuilding the affected turns. They also rewrite instructions drafted from part captions to describe the intended part, position, and orientation clearly. Appendix~\ref{app:construction} describes the engine, the chain format and the review procedure.

\begin{figure}[t]
\centering
\setlength{\abovecaptionskip}{5pt}\setlength{\belowcaptionskip}{2pt}
\ehfit{\begin{tikzpicture}
\begin{axis}[name=fa, at={(0,0)}, width=4.5cm, height=3.3cm, eh-bar, title={(a) Edits by family},
  xbar stacked, bar width=5.6pt, y dir=reverse, ytick={0,...,5}, yticklabels={{R\,\textcolor{pl-mute}{remove only}},{P\,\textcolor{pl-mute}{replace only}},{M\,\textcolor{pl-mute}{material}},{A\,\textcolor{pl-mute}{additions}},{G\,\textcolor{pl-mute}{geometry mix}},{X\,\textcolor{pl-mute}{geometry + material}}}, xmin=0, xmax=860, axis x line*=bottom,
  xtick={0,400,800}, xticklabels={0,400,800}, enlarge y limits=0.1, yticklabel style={anchor=east}, ytick pos=left, ytick style={draw=none}, axis y line*=left, y axis line style={draw=none}]
\addplot[xbar stacked, fill=op-add, draw=none] coordinates {(0,0) (0,1) (0,2) (690,3) (174,4) (104,5)};
\addplot[xbar stacked, fill=op-rem, draw=none] coordinates {(192,0) (0,1) (0,2) (59,3) (172,4) (87,5)};
\addplot[xbar stacked, fill=op-rep, draw=none] coordinates {(0,0) (293,1) (0,2) (0,3) (226,4) (110,5)};
\addplot[xbar stacked, fill=op-tex, draw=none] coordinates {(0,0) (0,1) (429,2) (0,3) (0,4) (219,5)};
\node[anchor=west, font=\sffamily\fontsize{6}{7}\selectfont, text=pl-mute] at (axis cs:192,0) {\,192};
\node[anchor=west, font=\sffamily\fontsize{6}{7}\selectfont, text=pl-mute] at (axis cs:293,1) {\,293};
\node[anchor=west, font=\sffamily\fontsize{6}{7}\selectfont, text=pl-mute] at (axis cs:429,2) {\,429};
\node[anchor=west, font=\sffamily\fontsize{6}{7}\selectfont, text=pl-mute] at (axis cs:749,3) {\,749};
\node[anchor=west, font=\sffamily\fontsize{6}{7}\selectfont, text=pl-mute] at (axis cs:572,4) {\,572};
\node[anchor=west, font=\sffamily\fontsize{6}{7}\selectfont, text=pl-mute] at (axis cs:520,5) {\,520};
\end{axis}
\begin{axis}[name=fb, at={($(fa.east)+(0.75cm,0)$)}, anchor=west, width=4.0cm, height=3.3cm, eh-bar, title={(b) Chain length (turns)},
  ybar, bar width=3.4pt, xmin=2.3, xmax=19.7, ymin=0, ymax=125, xtick={3,6,10,14,19}, xticklabels={3,6,10,14,19+}, ytick={0,50,100}, yticklabels={0,50,100}, axis x line*=bottom, axis y line*=left]
\addplot[fill=eh-main, draw=none] coordinates {(3,42) (4,79) (5,73) (6,111) (7,64) (8,45) (9,13) (10,17) (11,6) (12,1) (13,0) (14,1) (15,2) (16,1) (17,0) (18,0) (19,2)};
\draw[pl-ink, densely dashed, line width=0.5pt] (axis cs:6,0) -- (axis cs:6,118) node[above, font=\sffamily\fontsize{6}{7}\selectfont, text=pl-ink] {median 6};
\node[anchor=east, font=\sffamily\fontsize{6}{7}\selectfont, text=pl-mute, align=right] at (axis cs:19.6,62) {longest: 30};
\end{axis}
\begin{axis}[name=fc, at={($(fb.east)+(0.7cm,0)$)}, anchor=west, width=4.1cm, height=3.3cm, eh-bar, title={(c) Operation mix by turn},
  ybar stacked, bar width=6.4pt, xmin=0.4, xmax=10.6, ymin=0, ymax=100, xtick={1,3,5,7,10}, xticklabels={1,3,5,7,10+}, ytick={0,50,100},
  yticklabels={0,50,100\%}, axis x line*=bottom, axis y line*=left]
\addplot[ybar stacked, fill=op-add, draw=white, line width=0.3pt] coordinates {(1,34.6) (2,39.4) (3,37.9) (4,36.4) (5,32.4) (6,27.8) (7,34.9) (8,37.5) (9,23.3) (10,32.2)};
\addplot[ybar stacked, fill=op-rem, draw=white, line width=0.3pt] coordinates {(1,15.8) (2,16.4) (3,19.7) (4,18.3) (5,20.2) (6,23.2) (7,19.1) (8,11.4) (9,30.2) (10,18.4)};
\addplot[ybar stacked, fill=op-rep, draw=white, line width=0.3pt] coordinates {(1,30.2) (2,21.4) (3,22.1) (4,21.9) (5,22.6) (6,22.1) (7,19.1) (8,21.6) (9,11.6) (10,16.1)};
\addplot[ybar stacked, fill=op-tex, draw=white, line width=0.3pt] coordinates {(1,19.5) (2,22.8) (3,20.4) (4,23.4) (5,24.7) (6,27.0) (7,27.0) (8,29.5) (9,34.9) (10,33.3)};
\end{axis}
\begin{axis}[name=fd, at={($(fc.east)+(1.5cm,0)$)}, anchor=west, width=2.3cm, height=3.3cm, eh-bar, title={(d) Hosts},
  xbar, bar width=5.6pt, y dir=reverse, ytick={0,...,5}, yticklabels={{characters},{robots},{animals},{vehicles},{objects},{scenes}}, xmin=0, xmax=78, axis x line*=bottom, xtick={0,50}, xticklabels={0,50},
  enlarge y limits=0.12, axis y line*=left, y axis line style={draw=none}, ytick style={draw=none}]
\addplot[fill=eh-main, draw=none] coordinates {(48,0) (41,1) (30,2) (39,3) (61,4) (33,5)};
\node[anchor=west, font=\sffamily\fontsize{6}{7}\selectfont, text=pl-mute] at (axis cs:48,0) {\,48};
\node[anchor=west, font=\sffamily\fontsize{6}{7}\selectfont, text=pl-mute] at (axis cs:41,1) {\,41};
\node[anchor=west, font=\sffamily\fontsize{6}{7}\selectfont, text=pl-mute] at (axis cs:30,2) {\,30};
\node[anchor=west, font=\sffamily\fontsize{6}{7}\selectfont, text=pl-mute] at (axis cs:39,3) {\,39};
\node[anchor=west, font=\sffamily\fontsize{6}{7}\selectfont, text=pl-mute] at (axis cs:61,4) {\,61};
\node[anchor=west, font=\sffamily\fontsize{6}{7}\selectfont, text=pl-mute] at (axis cs:33,5) {\,33};
\end{axis}
\path let \p1 = (current bounding box.south west), \p2 = (current bounding box.south east) in node[anchor=north, font=\sffamily\fontsize{6.4}{7.4}\selectfont, inner sep=0pt] at ({(\x1+\x2)/2}, {\y1 - 0.25cm}) {\textcolor{op-add}{\scalebox{1.7}{$\bullet$}}\,add 968 \hspace{14pt} \textcolor{op-rem}{\scalebox{1.7}{$\bullet$}}\,remove 510 \hspace{14pt} \textcolor{op-rep}{\scalebox{1.7}{$\bullet$}}\,replace 629 \hspace{14pt} \textcolor{op-tex}{\scalebox{1.7}{$\bullet$}}\,retexture 648};
\end{tikzpicture}}
\caption{\textbf{EditHero statistics.} The benchmark contains \nChains{} chains, \nEdits{} edits, and \nHosts{} hosts. (a) Edits by chain family and operation. (b) Chain lengths, with lengths 19 and above pooled. (c) Share of each operation by turn index, with turns 10 and above pooled. (d) Host categories.}
\label{fig:stats}
\end{figure}
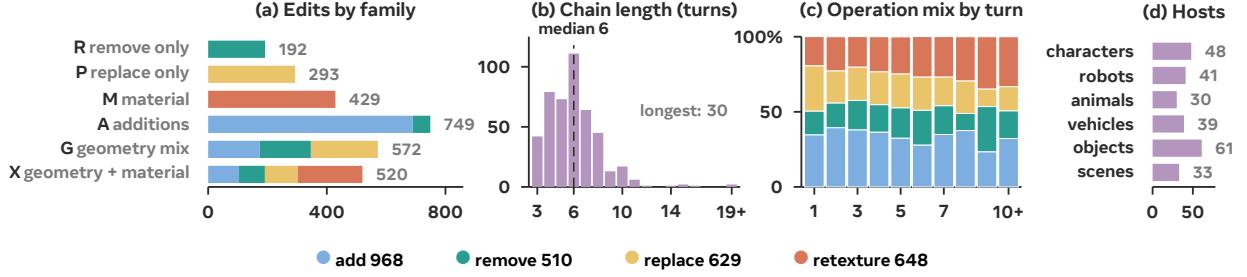

\noindent\textbf{Statistics.}
EditHero contains \nChains{} chains and \nEdits{} edits across \nHosts{} hosts (Figure~\ref{fig:stats}). Chains have \minTurns{} to \maxTurns{} turns, with a median of 6, and the median instruction is 17 words long. Six families group chains by their allowed operations. Removal (R), replacement (P), and material (M) chains contain only their respective operations. Addition (A) chains primarily add parts. Geometry (G) chains combine additions, removals, and replacements, while geometry-and-material (X) chains also include retexturing. Appendix Table~\ref{tab:bench-compare} compares EditHero with existing benchmarks.
\section{Evaluation Protocol}
\label{sec:protocol}

\noindent\textbf{Setting.}
We evaluate the editing methods under \emph{self-rollout}, in which turn $k$ takes the method's own output from turn $k-1$ as input. At each turn a method receives its current state, the instruction and the target render, and VoxHammer also receives a ground-truth edit mask. Ground-truth states are used only for scoring. All states of a chain share one coordinate frame and 4 fixed camera views, 1 conditioning view and 3 held-out views. Image metrics are averaged over the 4 views, and Appendix~\ref{app:views} reports the conditioning view and the held-out views separately. Scores are averaged over the turns a method completes. Appendix~\ref{app:protocol} provides details.

\noindent\textbf{Design of metrics.}
Previous 3D editing evaluations follow 2 designs. When a target asset exists, they compare the edited object with it as a whole, using Chamfer distance, normal consistency, F-score and image similarity of the renders~\citep{editformer,partflow,ma2025steer3d,preintner2026threedmorph}. Without a target asset, they measure how well the unedited region keeps the source object and judge the edit by its similarity to an edited image or to the text~\citep{voxhammer,nano3d,zhou2026anchorflow,gat2026shapeup,nie2026editflow3d,patro2026editbench3d}. Both designs score a single edit applied to a clean source object. In our chains, the input to each turn is the method's own previous output, and every turn still has an exact target. Whole-object F-score can be a poor indicator of editing ability. Because each turn changes only a small region, simply returning the input already yields a high score. We therefore evaluate the edit region and the unchanged region separately. In addition, under self-rollout the method's own states drift from the ground truth starting from the first turn. To isolate the effect of the current edit, we compare each output against the method's own previous output rather than against the ground-truth previous state.

\noindent\textbf{Two reference baselines.}
We use 2 baselines as reference points at opposite ends. The \emph{no-operation (no-op) baseline} returns the initial object $S_0$ at every turn. It keeps everything and edits nothing, so a metric on which it scores well rewards preservation rather than editing. The \emph{regeneration baseline} runs TRELLIS~\citep{trellis} on the target render of each turn, without the instruction or any previous state, and aligns the result to the ground-truth state of that turn by a similarity transform. It sees the target at every turn but keeps nothing from the previous output. A good editor should follow instructions better than the no-op baseline and keep the unedited parts better than regeneration. We also use the no-op baseline to check every metric, because a metric on which doing nothing scores well cannot measure editing.

\noindent\textbf{Regions and region-restricted metrics.}
We define the edit region from the parts named by the instruction in the ground-truth states before and after turn $k$. Its complement is the unchanged region. Let $\eparts{k}$ denote these parts and $\restr{S}{\eparts{k}}$ their geometry in state $S$, empty if they are absent. The union of their regions before and after the edit covers additions, removals, replacements, and retextures under one definition.

For 3D evaluation, let $\vox{\cdot}$ voxelize a surface in the chain's grid $\grid=\{0,\dots,63\}^3$, and let $\dil{\cdot}$ dilate a set by one cell, clamped to the grid. We define
\begin{equation}
\editR{k}=\dil{\vox{\restr{\gtS{k-1}}{\eparts{k}}}\cup\vox{\restr{\gtS{k}}{\eparts{k}}}},\qquad
\keepR{k}=\grid\setminus\editR{k}.
\label{eq:regions3d}
\end{equation}
For a fixed camera view, let $\proj{S}{\eparts{k}}$ be the depth-tested projection of the named parts, excluding pixels occluded by other geometry. With one-pixel dilation over the image domain $\imgdom$, the corresponding regions are
\begin{equation}
\editRimg{k}=\dil{\proj{\gtS{k-1}}{\eparts{k}}\cup\proj{\gtS{k}}{\eparts{k}}},\qquad
\keepRimg{k}=\imgdom\setminus\editRimg{k}.
\label{eq:regions2d}
\end{equation}
These regions depend on part identity rather than image or voxel differences. Thus, lighting changes and retextures do not redefine where an edit is allowed, while unintended changes outside the named parts fall in the unchanged region.
We then write $m_R(\cdot,\cdot)$ for a metric evaluated only in region $R$ of Eq.~\eqref{eq:regions3d} or~\eqref{eq:regions2d}. For 3D states $X$ and $Y$, for example,
\begin{equation}
\IoUon{R}{X}{Y}=\frac{\lvert\vox{X}\cap\vox{Y}\cap R\rvert}{\lvert(\vox{X}\cup\vox{Y})\cap R\rvert}.
\label{eq:ops3d}
\end{equation}
Surface samples are assigned to regions by their voxels, and image metrics aggregate per-pixel values within the specified region. We also evaluate the whole object by setting $R=\grid$ or $R=\imgdom$. We compare each output $\predS{k}$ with its target $\gtS{k}$ in the edit region, the unchanged region and the whole object, using the fixed-view renders for the 2D comparisons. Table~\ref{tab:metrics} summarizes the operators, and Appendix~\ref{app:metrics} defines them in full.

\noindent\textbf{Instruction following.}
Target similarity in the edit region can be high simply because the named part was already there. We therefore also measure \emph{instruction following} (IF) against the method's own previous output. Under self-rollout this output can already differ from the ground truth, so IF counts only the change still required from it. Otherwise a part lost at an earlier turn would count as a successful removal now, and the no-op baseline would get credit for edits it never made. For additions, $A_k$ contains required new voxels still missing before the turn. For removals, $D_k$ contains voxels that must disappear but remain in the previous output:
\begin{equation}
\begin{aligned}
\mathrm{IF}^{+}_k&=\Recallon{A_k}{\predS{k}}{\gtS{k}}, &
A_k&=\bigl(\vox{\gtS{k}}\setminus\vox{\gtS{k-1}}\bigr)\setminus\vox{\predS{k-1}}\subseteq\editR{k},\\
\mathrm{IF}^{-}_k&=1-\Recallon{D_k}{\predS{k}}{\gtS{k-1}}, &
D_k&=\bigl(\vox{\gtS{k-1}}\setminus\vox{\gtS{k}}\bigr)\cap\vox{\predS{k-1}}\subseteq\editR{k}.
\end{aligned}
\label{eq:if}
\end{equation}
IF is IF$^{+}$ for additions, IF$^{-}$ for removals, and their harmonic mean for replacements. It is undefined when the relevant region is too small and for retextures, which require no geometric change. Appendix Table~\ref{tab:if-counts} reports how many turns have a defined IF.

\noindent\textbf{Content consistency.}
Outside the edit region we measure what the current turn keeps from the method's own previous output ($\mathrm{CC}^{\text{prev}}$) and what is left of the initial host ($\mathrm{CC}^{0}$):
\begin{equation}
\mathrm{CC}^{\text{prev}}_k=\IoUon{\keepR{k}}{\predS{k}}{\predS{k-1}},\qquad
\mathrm{CC}^{0}_k=\IoUon{\keepRcum{k}}{\predS{k}}{\gtS{0}},\quad
\keepRcum{k}=\grid\setminus\textstyle\bigcup_{j\le k}\editR{j}.
\label{eq:cc}
\end{equation}
The cumulative region excludes the named parts before and after every edit so far. By construction, the no-op baseline has IF $=0$ and CC $=1$. IF therefore cannot be won by doing nothing, and CC has to be read together with IF. Appendix~\ref{app:protocol} illustrates these choices with examples.

\begin{table}[t]
\centering\footnotesize\setlength{\tabcolsep}{4pt}
\caption{\textbf{Region-restricted evaluation metrics.} $X,Y$ are output and reference states; $I,J$ are their fixed-view renders; and $R$ is a region from Eq.~\eqref{eq:regions3d} or~\eqref{eq:regions2d}. Full definitions appear in Appendix~\ref{app:metrics}.}
\label{tab:metrics}
\ehsep{10pt}\ehfitplain{\begin{tabular}{lll@{\hspace{8pt}}lll}
\toprule
\multicolumn{3}{c}{3D} & \multicolumn{3}{c}{2D} \\
\cmidrule(r){1-3}\cmidrule(l){4-6}
$\IoUon{R}{X}{Y}$ & voxel IoU & $\uparrow$ & $\PSNRon{R}{I}{J}$ & peak signal-to-noise ratio & $\uparrow$ \\
$\Recallon{R}{X}{Y}$ & voxel recall of $Y$ & $\uparrow$ & $\SSIMon{R}{I}{J}$ & SSIM map~\citep{ssim} & $\uparrow$ \\
$\CDon{R}{X}{Y}$ & Chamfer distance & $\downarrow$ & $\LPIPSon{R}{I}{J}$ & LPIPS map~\citep{lpips} & $\downarrow$ \\
$\Fon{R}{X}{Y}$ & F-score at distance $\tau$ & $\uparrow$ & $\DINOon{R}{I}{J}$ & DINOv2 cosine~\citep{oquab2023dinov2} & $\uparrow$ \\
 & & & $\MAEon{R}{I}{J}$ & mean absolute error & $\downarrow$ \\
\bottomrule
\end{tabular}}
\end{table}
\section{Experiments}
\label{sec:experiments}

\subsection{Setup}
We evaluate 4 non-agentic 3D methods, PartFlow~\citep{partflow}, Nano3D~\citep{nano3d}, 3DEditFormer~\citep{editformer} and VoxHammer~\citep{voxhammer}, with their released code, weights and sampling settings. Each of these 4 methods runs under self-rollout on all \nEvalChains{} chains (\nEvalEdits{} turns) and receives, at each turn, its own previous output, the instruction and the target render in the fixed view of the chain. VoxHammer also receives the ground-truth edit mask. Nano3D has no material mode, so its material turns run in its replace mode. We also run the no-op and regeneration baselines of Section~\ref{sec:protocol}. Appendix~\ref{app:setup} describes the method settings and execution coverage.

\subsection{Main results}
\label{sec:analysis}
Table~\ref{tab:regions} reports geometry and image metrics in the edit region, the unchanged region and the whole object. The 2 baselines appear above the non-agentic methods.

\begin{table}[t]
\centering\footnotesize\setlength{\tabcolsep}{3.2pt}
\caption{\textbf{Comparison of non-agentic methods and baselines.} Self-rollout on \nEvalChains{} chains; means over completed turns with a defined score. IF: instruction following against the method's own previous output. CC$^{\text{prev}}$: IoU of the unchanged region to the method's own previous output. CD: Chamfer distance ($\times 100$). Drift: mean absolute change of the unchanged image region against the method's own previous render. Best non-agentic method in bold.}
\label{tab:regions}
\ehsep{4pt}\ehfit{\begin{tabular}{l cc cc cc c cc cc}
\toprule
 & \multicolumn{6}{c}{3D} & \multicolumn{5}{c}{2D} \\ \cmidrule(lr){2-7}\cmidrule(lr){8-12}
 & \multicolumn{2}{c}{edit $\editR{k}$} & \multicolumn{2}{c}{unchanged $\keepR{k}$} & \multicolumn{2}{c}{whole} & edit & \multicolumn{2}{c}{unchanged} & \multicolumn{2}{c}{whole} \\ \cmidrule(lr){2-3}\cmidrule(lr){4-5}\cmidrule(lr){6-7}\cmidrule(lr){8-8}\cmidrule(lr){9-10}\cmidrule(lr){11-12}
Method & IF$\uparrow$ & F$\uparrow$ & CC$^{\text{prev}}\uparrow$ & F$\uparrow$ & F$\uparrow$ & CD$\downarrow$ & LPIPS$\downarrow$ & LPIPS$\downarrow$ & drift$\downarrow$ & LPIPS$\downarrow$ & PSNR$\uparrow$ \\
\midrule
No-op baseline & 0.00 & 0.56 & 1.00 & 0.82 & 0.77 & 2.03 & 0.569 & 0.130 & 0.000 & 0.151 & 18.7 \\
TRELLIS regen. & 0.50 & 0.43 & 0.38 & 0.42 & 0.43 & 2.07 & 0.402 & 0.193 & 0.041 & 0.203 & 15.8 \\
\midrule
PartFlow & 0.12 & 0.47 & 0.86 & 0.71 & 0.67 & 2.15 & 0.516 & 0.166 & 0.014 & 0.181 & 17.0 \\
Nano3D & 0.31 & \textbf{0.56} & \textbf{0.89} & \textbf{0.75} & \textbf{0.71} & \textbf{1.99} & \textbf{0.474} & \textbf{0.147} & \textbf{0.009} & \textbf{0.163} & \textbf{17.7} \\
3DEditFormer & 0.20 & 0.43 & 0.77 & 0.57 & 0.55 & 2.80 & 0.510 & 0.188 & 0.018 & 0.203 & 16.2 \\
VoxHammer & \textbf{0.32} & 0.38 & 0.71 & 0.55 & 0.52 & 2.69 & 0.521 & 0.205 & 0.022 & 0.219 & 15.2 \\
\bottomrule
\end{tabular}
}
\end{table}

\noindent \textbf{Whole-object scores favor doing nothing.}
On the whole object, the no-op baseline beats every non-agentic method on F-score, LPIPS and PSNR. A single turn changes only a small part of the object, so returning the input already scores well. Only the edit region tells real edits apart from doing nothing. There, every method is closer to the target than the no-op baseline (LPIPS 0.474--0.521 against 0.569). The regeneration baseline sits at the other extreme. It rebuilds the whole object from the target render at every turn, so it has the highest IF (0.50), but its unchanged-region IoU with its own previous output is only 0.38. A good editor needs both high IF and good preservation, and no non-agentic method reaches regeneration on IF or the no-op baseline on preservation.

\noindent \textbf{Instructions are followed poorly from the first turn.}
IF is only 0.12--0.32 over all turns, and it is already low at turn~1 (0.24--0.36), before any error can accumulate. Nano3D and VoxHammer follow instructions best, but VoxHammer keeps much less of the rest of the object (unchanged-region IoU 0.71 against 0.89).

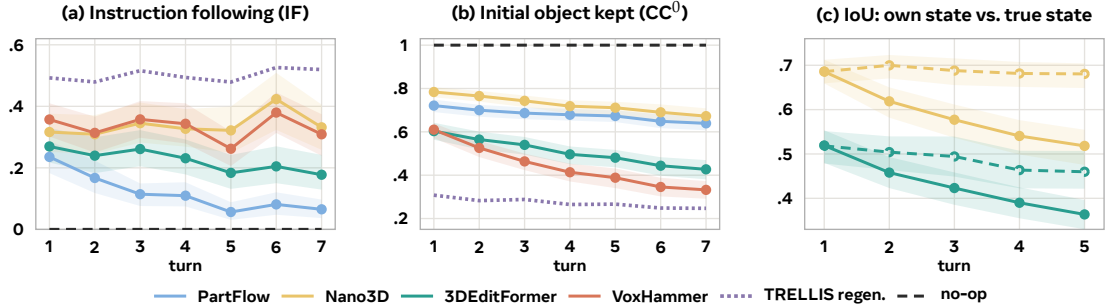
\begin{figure}[t]
\centering
\ehfitplain{\begin{tikzpicture}
\begin{axis}[name=pif_self, at={(0,0)}, font=\sffamily\fontsize{5.6}{6.4}\selectfont, title style={font=\sffamily\bfseries\fontsize{6.4}{7}\selectfont, yshift=-5pt}, width=4.75cm, height=3.6cm, xmin=0.7, xmax=7.3, xtick={1,...,7}, xticklabels={1,2,3,4,5,6,7}, xlabel={turn}, xlabel style={yshift=4pt}, axis line style={draw=pl-mute!60}, tick style={draw=none}, grid=major, grid style={pl-line!70}, every axis plot/.append style={line width=0.9pt, mark size=1.1pt}, title={(a) Instruction following (IF)}, ytick={0,0.2,0.4,0.6}, yticklabels={0,.2,.4,.6}, ymin=0, ymax=0.62]
\addplot[name path=lo1,draw=none,forget plot] table[x=turn,y=lo]{figures/turns/if_self_partflow_reference_fixed_chain_v2_cam2.dat};\addplot[name path=hi1,draw=none,forget plot] table[x=turn,y=hi]{figures/turns/if_self_partflow_reference_fixed_chain_v2_cam2.dat};\addplot[c1,opacity=0.13,forget plot] fill between[of=lo1 and hi1];
\addplot[c1,mark=*] table[x=turn,y=mean]{figures/turns/if_self_partflow_reference_fixed_chain_v2_cam2.dat};

\addplot[name path=lo2,draw=none,forget plot] table[x=turn,y=lo]{figures/turns/if_self_nano_reference_fixed_chain_v2_cam2.dat};\addplot[name path=hi2,draw=none,forget plot] table[x=turn,y=hi]{figures/turns/if_self_nano_reference_fixed_chain_v2_cam2.dat};\addplot[c2,opacity=0.13,forget plot] fill between[of=lo2 and hi2];
\addplot[c2,mark=*] table[x=turn,y=mean]{figures/turns/if_self_nano_reference_fixed_chain_v2_cam2.dat};

\addplot[name path=lo3,draw=none,forget plot] table[x=turn,y=lo]{figures/turns/if_self_3deditformer_reference_fixed_chain_v2_cam2.dat};\addplot[name path=hi3,draw=none,forget plot] table[x=turn,y=hi]{figures/turns/if_self_3deditformer_reference_fixed_chain_v2_cam2.dat};\addplot[c3,opacity=0.13,forget plot] fill between[of=lo3 and hi3];
\addplot[c3,mark=*] table[x=turn,y=mean]{figures/turns/if_self_3deditformer_reference_fixed_chain_v2_cam2.dat};

\addplot[name path=lo4,draw=none,forget plot] table[x=turn,y=lo]{figures/turns/if_self_voxhammer_reference_gtmask_fixed_chain_v2_cam2.dat};\addplot[name path=hi4,draw=none,forget plot] table[x=turn,y=hi]{figures/turns/if_self_voxhammer_reference_gtmask_fixed_chain_v2_cam2.dat};\addplot[c4,opacity=0.13,forget plot] fill between[of=lo4 and hi4];
\addplot[c4,mark=*] table[x=turn,y=mean]{figures/turns/if_self_voxhammer_reference_gtmask_fixed_chain_v2_cam2.dat};

\addplot[c5, densely dotted, mark=none, line width=1.1pt] table[x=turn,y=mean]{figures/turns/if_self_trellis_regen_fixed_chain_v2_cam2.dat};
\addplot[pl-ink, densely dashed, mark=none] table[x=turn,y=mean]{figures/turns/if_self_noop.dat};
\end{axis}
\begin{axis}[name=pcc_turn0_dil, at={($(pif_self.east)+(0.9cm,0)$)}, anchor=west, font=\sffamily\fontsize{5.6}{6.4}\selectfont, title style={font=\sffamily\bfseries\fontsize{6.4}{7}\selectfont, yshift=-5pt}, width=4.75cm, height=3.6cm, xmin=0.7, xmax=7.3, xtick={1,...,7}, xticklabels={1,2,3,4,5,6,7}, xlabel={turn}, xlabel style={yshift=4pt}, axis line style={draw=pl-mute!60}, tick style={draw=none}, grid=major, grid style={pl-line!70}, every axis plot/.append style={line width=0.9pt, mark size=1.1pt}, title={(b) Initial object kept (CC$^{0}$)}, ytick={0.2,0.4,0.6,0.8,1}, yticklabels={.2,.4,.6,.8,1}, ymin=0.15, ymax=1.03]
\addplot[name path=lo7,draw=none,forget plot] table[x=turn,y=lo]{figures/turns/cc_turn0_dil_partflow_reference_fixed_chain_v2_cam2.dat};\addplot[name path=hi7,draw=none,forget plot] table[x=turn,y=hi]{figures/turns/cc_turn0_dil_partflow_reference_fixed_chain_v2_cam2.dat};\addplot[c1,opacity=0.13,forget plot] fill between[of=lo7 and hi7];
\addplot[c1,mark=*] table[x=turn,y=mean]{figures/turns/cc_turn0_dil_partflow_reference_fixed_chain_v2_cam2.dat};

\addplot[name path=lo8,draw=none,forget plot] table[x=turn,y=lo]{figures/turns/cc_turn0_dil_nano_reference_fixed_chain_v2_cam2.dat};\addplot[name path=hi8,draw=none,forget plot] table[x=turn,y=hi]{figures/turns/cc_turn0_dil_nano_reference_fixed_chain_v2_cam2.dat};\addplot[c2,opacity=0.13,forget plot] fill between[of=lo8 and hi8];
\addplot[c2,mark=*] table[x=turn,y=mean]{figures/turns/cc_turn0_dil_nano_reference_fixed_chain_v2_cam2.dat};

\addplot[name path=lo9,draw=none,forget plot] table[x=turn,y=lo]{figures/turns/cc_turn0_dil_3deditformer_reference_fixed_chain_v2_cam2.dat};\addplot[name path=hi9,draw=none,forget plot] table[x=turn,y=hi]{figures/turns/cc_turn0_dil_3deditformer_reference_fixed_chain_v2_cam2.dat};\addplot[c3,opacity=0.13,forget plot] fill between[of=lo9 and hi9];
\addplot[c3,mark=*] table[x=turn,y=mean]{figures/turns/cc_turn0_dil_3deditformer_reference_fixed_chain_v2_cam2.dat};

\addplot[name path=lo10,draw=none,forget plot] table[x=turn,y=lo]{figures/turns/cc_turn0_dil_voxhammer_reference_gtmask_fixed_chain_v2_cam2.dat};\addplot[name path=hi10,draw=none,forget plot] table[x=turn,y=hi]{figures/turns/cc_turn0_dil_voxhammer_reference_gtmask_fixed_chain_v2_cam2.dat};\addplot[c4,opacity=0.13,forget plot] fill between[of=lo10 and hi10];
\addplot[c4,mark=*] table[x=turn,y=mean]{figures/turns/cc_turn0_dil_voxhammer_reference_gtmask_fixed_chain_v2_cam2.dat};

\addplot[c5, densely dotted, mark=none, line width=1.1pt] table[x=turn,y=mean]{figures/turns/cc_turn0_dil_trellis_regen_fixed_chain_v2_cam2.dat};
\addplot[pl-ink, densely dashed, mark=none] table[x=turn,y=mean]{figures/turns/cc_turn0_dil_noop.dat};
\end{axis}
\begin{axis}[name=pgt, at={($(pcc_turn0_dil.east)+(0.9cm,0)$)}, anchor=west, font=\sffamily\fontsize{5.6}{6.4}\selectfont, title style={font=\sffamily\bfseries\fontsize{6.4}{7}\selectfont, yshift=-5pt}, width=4.75cm, height=3.6cm, xmin=0.7, xmax=7.3, xtick={1,...,7}, xticklabels={1,2,3,4,5,6,7}, xlabel={turn}, xlabel style={yshift=4pt}, axis line style={draw=pl-mute!60}, tick style={draw=none}, grid=major, grid style={pl-line!70}, every axis plot/.append style={line width=0.9pt, mark size=1.1pt}, xmax=5.3, xtick={1,...,5}, xticklabels={1,2,3,4,5}, title={(c) IoU: own state vs.\ true state}, ytick={0.4,0.5,0.6,0.7}, yticklabels={.4,.5,.6,.7}, ymin=0.33, ymax=0.76]
\addplot[name path=lo13,draw=none,forget plot] table[x=turn,y=lo]{figures/turns/iou_own_nano_reference_fixed_chain_v2_cam2.dat};\addplot[name path=hi13,draw=none,forget plot] table[x=turn,y=hi]{figures/turns/iou_own_nano_reference_fixed_chain_v2_cam2.dat};\addplot[c2,opacity=0.13,forget plot] fill between[of=lo13 and hi13];
\addplot[c2,mark=*] table[x=turn,y=mean]{figures/turns/iou_own_nano_reference_fixed_chain_v2_cam2.dat};

\addplot[name path=lo14,draw=none,forget plot] table[x=turn,y=lo]{figures/turns/iou_reset_nano_reference_fixed_chain_v2_cam2.dat};\addplot[name path=hi14,draw=none,forget plot] table[x=turn,y=hi]{figures/turns/iou_reset_nano_reference_fixed_chain_v2_cam2.dat};\addplot[c2,opacity=0.13,forget plot] fill between[of=lo14 and hi14];
\addplot[c2,densely dashed, mark=o] table[x=turn,y=mean]{figures/turns/iou_reset_nano_reference_fixed_chain_v2_cam2.dat};

\addplot[name path=lo15,draw=none,forget plot] table[x=turn,y=lo]{figures/turns/iou_own_3deditformer_reference_fixed_chain_v2_cam2.dat};\addplot[name path=hi15,draw=none,forget plot] table[x=turn,y=hi]{figures/turns/iou_own_3deditformer_reference_fixed_chain_v2_cam2.dat};\addplot[c3,opacity=0.13,forget plot] fill between[of=lo15 and hi15];
\addplot[c3,mark=*] table[x=turn,y=mean]{figures/turns/iou_own_3deditformer_reference_fixed_chain_v2_cam2.dat};

\addplot[name path=lo16,draw=none,forget plot] table[x=turn,y=lo]{figures/turns/iou_reset_3deditformer_reference_fixed_chain_v2_cam2.dat};\addplot[name path=hi16,draw=none,forget plot] table[x=turn,y=hi]{figures/turns/iou_reset_3deditformer_reference_fixed_chain_v2_cam2.dat};\addplot[c3,opacity=0.13,forget plot] fill between[of=lo16 and hi16];
\addplot[c3,densely dashed, mark=o] table[x=turn,y=mean]{figures/turns/iou_reset_3deditformer_reference_fixed_chain_v2_cam2.dat};

\end{axis}
\matrix[anchor=north, font=\sffamily\fontsize{5.6}{6}\selectfont, column sep=3pt, inner sep=0pt] at ($(pif_self.south west)!0.5!(pgt.south east)+(0,-0.62cm)$) { \draw[c1, line width=0.9pt] (0,0) -- (0.34,0); & \node[anchor=west, inner sep=0]{PartFlow}; & \draw[c2, line width=0.9pt] (0,0) -- (0.34,0); & \node[anchor=west, inner sep=0]{Nano3D}; & \draw[c3, line width=0.9pt] (0,0) -- (0.34,0); & \node[anchor=west, inner sep=0]{3DEditFormer}; & \draw[c4, line width=0.9pt] (0,0) -- (0.34,0); & \node[anchor=west, inner sep=0]{VoxHammer}; & \draw[c5, densely dotted, line width=1.1pt] (0,0) -- (0.34,0); & \node[anchor=west, inner sep=0]{TRELLIS regen.}; & \draw[pl-ink, densely dashed, line width=0.9pt] (0,0) -- (0.34,0); & \node[anchor=west, inner sep=0]{no-op}; \\ };
\end{tikzpicture}}
\caption{\textbf{Along a chain.} (a,b) Means on the same \turnFigChains{} chains with at least 7 turns, completed by all 4 non-agentic methods (\turnFigHosts{} hosts). Only defined scores enter each mean; shading shows 95\% host-clustered bootstrap intervals. (a)~IF relative to the method's previous output. (b)~CC$^0$ in regions never edited so far. (c)~IoU to the target on the \gtResetFixedChains{} chains of the reset diagnostic with at least 5 turns (Appendix~\ref{app:gtreset}). Solid lines: each turn starts from the model's own output. Dashed lines: each turn starts from the ground-truth previous state.}
\label{fig:turns}
\end{figure}

\noindent \textbf{Errors accumulate under self-rollout.}
Along a chain, IF stays low and all 4 methods keep losing the initial object (Figure~\ref{fig:turns}a,b). From turn~1 to turn~7, CC$^{0}$ falls from 0.60--0.78 to 0.33--0.67, and VoxHammer loses the most. The reset diagnostic suggests that much of this loss is inherited, because each turn starts from the unintended changes of the previous one. When every turn instead starts from the ground-truth previous state, IoU to the target at turn~5 is 0.68 for Nano3D and 0.46 for 3DEditFormer, but under self-rollout it is only 0.52 and 0.36 (Figure~\ref{fig:turns}c). Appendices~\ref{app:longhorizon} and~\ref{app:material} report more results.

\subsection{Qualitative results}
\begin{figure}[t]
\centering
\ehfit{\input{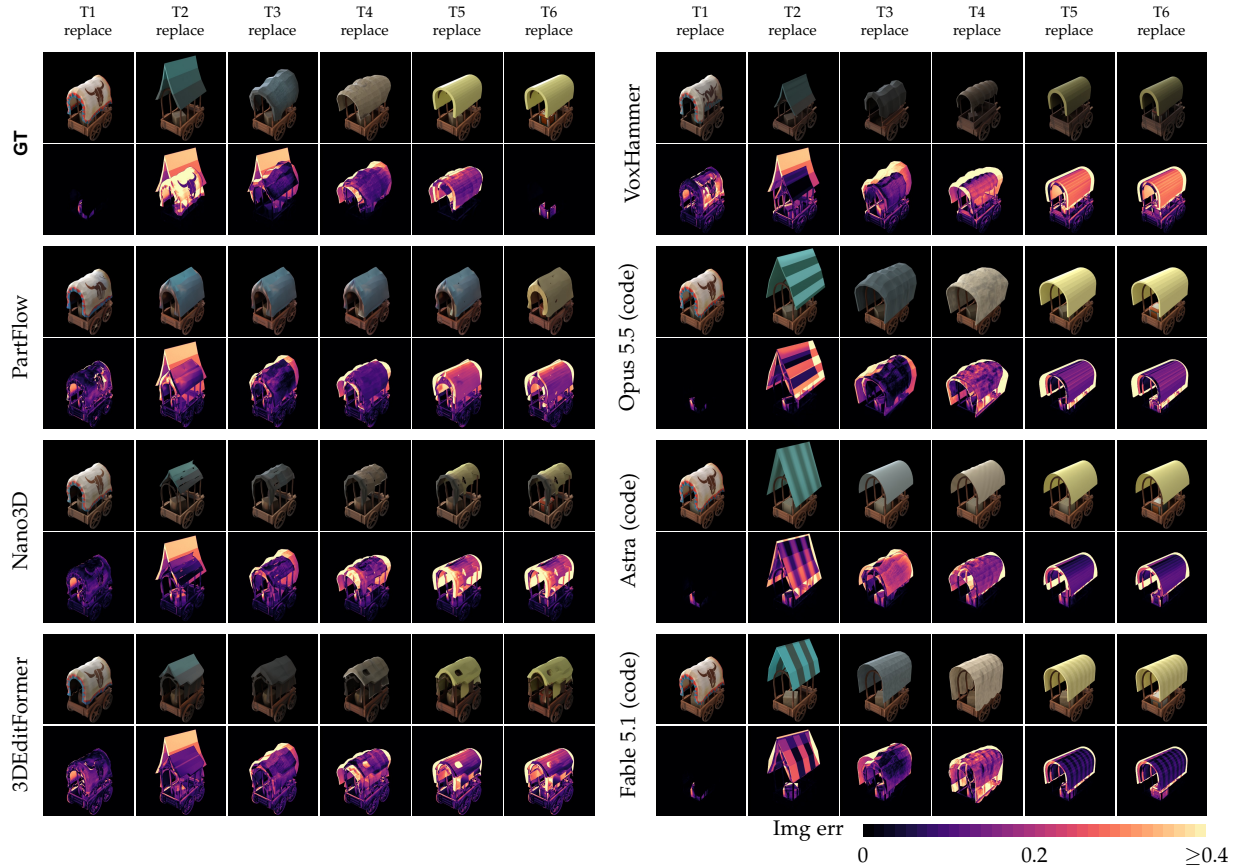}}
\caption{\textbf{Six consecutive replacements on a covered wagon} (turns 1--6). The barrel becomes a wooden crate, the cover becomes a striped peaked canopy and then gray, beige and pale yellow fabric covers, and the crate becomes a science-fiction storage box. Columns are turns. In each block, the upper row is the method's own output under self-rollout, and the lower row is its per-pixel mean absolute RGB error to the ground truth of that turn. In the ground-truth block, the lower row shows the change from the previous turn. Left: the ground truth and 3 non-agentic methods. Right: VoxHammer and 3 agents (Section~\ref{sec:agentic}). Same camera and lighting throughout.}
\label{fig:qual-webview}
\end{figure}

Figure~\ref{fig:qual-webview} follows 6 replacements on a covered wagon from the comparison subset, in which the cover is replaced 4 times in a row. No non-agentic method follows this sequence. PartFlow keeps a blue rounded cover from turn~2 to turn~5, and the other 3 keep a dark cover and end with a dull olive instead of pale yellow. 3DEditFormer and VoxHammer also change the color of the wagon bed, which no instruction touches. The 3 LLMs shown (Opus~5.5, Astra and Fable~5.1) produce covers closer to the targets, and their errors stay on the cover. Appendix Figure~\ref{fig:qual-character} and Appendix~\ref{app:qual-methods} show more chains.

\subsection{Comparison with LLM/VLM agents}
\label{sec:agentic}
An LLM agent inspects the current mesh, writes code to add, remove, replace or retexture parts, and checks rendered views before it commits an edit. We evaluate 6 models: GLM~5.3 Flash, DeepSeek~V4.1 Flash, Sol~6, Astra, Fable~5.1 and Opus~5.5. Each chain is assigned to an independent agent using the same editing prompt. New parts must be built through code; external part libraries and generative 3D or image models are not allowed. Appendix~\ref{app:llm} gives the setup.

We compare these agents with the 4 non-agentic methods and both baselines on a shared \emph{comparison subset} of \llmChains{} chains (\llmTurns{} turns). We selected chains with clearly visible edits that resemble an artist's revisions, covering all 6 families and a range of operation combinations. The editing methods use self-rollout. Table~\ref{tab:agentic} reports the comparison, and Appendix~\ref{app:qual-llm} shows additional examples.

\begin{table}[t]
\centering\footnotesize
\caption{\textbf{LLM/VLM agents, non-agentic methods and baselines.} All rows use the \llmChains{} chains (\llmTurns{} turns) of the comparison subset, so the non-agentic rows differ from Table~\ref{tab:regions}. \emph{Completed / Total}: delivered meshes out of all scheduled turns. Means over completed turns with a defined value. $E$: edit region; $W$: whole object or image. Best method in bold, excluding baselines.}
\label{tab:agentic}
\ehfit{\begin{tabular}{l r cc c cc c}
\toprule
 & & \multicolumn{3}{c}{3D} & \multicolumn{3}{c}{2D} \\ \cmidrule(lr){3-5}\cmidrule(lr){6-8}
Method & Completed / Total & IF$\uparrow$ & CC$^{\text{prev}}\uparrow$ & F$_W\uparrow$ & LPIPS$_E\downarrow$ & LPIPS$_W\downarrow$ & PSNR$_W\uparrow$ \\
\midrule
No-op baseline & 416/416 & 0.00 & 1.00 & 0.75 & 0.567 & 0.154 & 18.2 \\
TRELLIS regen. & 416/416 & 0.49 & 0.35 & 0.37 & 0.411 & 0.228 & 15.0 \\
\midrule
\multicolumn{8}{@{}l}{\itshape Non-agentic methods} \\
PartFlow & 416/416 & 0.13 & 0.87 & 0.65 & 0.501 & 0.184 & 16.8 \\
Nano3D & 416/416 & 0.34 & 0.90 & 0.70 & 0.448 & 0.161 & 17.6 \\
3DEditFormer & 416/416 & 0.21 & 0.81 & 0.59 & 0.472 & 0.191 & 16.5 \\
VoxHammer & 377/416 & 0.36 & 0.73 & 0.55 & 0.490 & 0.213 & 15.7 \\
\midrule
\multicolumn{8}{@{}l}{\itshape LLM/VLM agents} \\
GLM~5.3 Flash & 416/416 & 0.26 & 0.95 & 0.75 & 0.447 & 0.136 & 21.3 \\
DeepSeek~V4.1 Flash & 416/416 & 0.42 & 0.96 & 0.78 & 0.366 & 0.106 & 24.0 \\
Sol~6 (medium) & 416/416 & 0.42 & \textbf{0.98} & 0.79 & 0.343 & 0.093 & 24.6 \\
Astra (medium) & 416/416 & 0.53 & \textbf{0.98} & 0.81 & 0.287 & 0.077 & 25.6 \\
Fable~5.1 & 416/416 & 0.52 & \textbf{0.98} & 0.80 & 0.294 & 0.080 & 25.4 \\
Opus~5.5 & 416/416 & \textbf{0.62} & \textbf{0.98} & \textbf{0.82} & \textbf{0.238} & \textbf{0.065} & \textbf{26.6} \\
\bottomrule
\end{tabular}
}
\end{table}

\noindent\textbf{LLMs preserve more and usually follow instructions better.}
All 6 LLMs keep the rest of the object better than any non-agentic method (CC$^{\text{prev}}$ \llmGLMCCprev{}--\llmOpusCCprev{} against at most \llmGenCCmax{}). All except GLM~5.3 Flash also follow instructions better (IF up to \llmOpusIF{} for Opus~5.5, against at most \llmGenIFmax{}). This is precisely the advantage of the bottom-up approach: an agent modifies only the named part and leaves the remainder of the asset untouched by construction. Removal is the easiest operation, since code can simply delete a part (IF \llmGLMIFrem{}--\llmOpusIFrem{}). Additions and replacements are harder (IF \llmGLMIFadd{}--\llmOpusIFadd{} and \llmGLMIFrep{}--\llmOpusIFrep{}), because new geometry built from code often misses the target's size or shape, as on the wagon cover in Figure~\ref{fig:qual-webview}.

Unlike the non-agentic methods, every LLM also beats the no-op baseline on whole-object LPIPS (\llmGLMLPIPSW{} for GLM~5.3 Flash to \llmOpusLPIPSW{} for Opus~5.5, against \llmNoopLPIPSW{}), and all of them deliver a mesh at every turn.

\noindent\textbf{LLMs are slow.}
An LLM needs about 1.5--6 minutes and 6--16 model calls per edit (medians, Appendix~\ref{app:llm}), while a non-agentic method takes 20--50 seconds on a local GPU. The best LLMs edit well, but they are still too slow for interactive use.
\section{Conclusion}
\label{sec:conclusion}
EditHero tests whether a 3D editing method can perform successive edits on an asset over many turns. Because its chains are assembled from parts, the correct state after every turn is known exactly, so we can score the requested change and the rest of the object separately. The non-agentic methods, which regenerate the whole object at every turn, often miss the edit from the first turn and let the unedited parts drift as the chain grows. On whole-object F-score, LPIPS and PSNR they even fall behind simply returning the input. LLM agents, which change only the parts required by instructions, keep the remainder of the asset intact, and most of them also follow instructions more closely. It seems that the agentic approach is naturally suited to editing. However, building new geometry from code remains hard, and an edit that takes minutes is still too slow for interactive work. It remains an open question whether we can find a proper combination of agentic and conventional methods that is both fast and accurate. We will release the data synthesis engine and the edit chains for further 3D editing research.

\bibliography{references}
\bibliographystyle{assets/plainnat}

\clearpage\beginappendix
\section{Extended related work}
\label{app:related}

\paragraph{3D generation.}
An influential line of text-to-3D generation optimizes a 3D representation against a 2D diffusion prior~\citep{dreamfusion,magic3d,prolificdreamer}, with related approaches using multi-view diffusion~\citep{zero123,mvdream,wonder3d} and feed-forward reconstruction~\citep{lrm,instantmesh}. Current systems train a diffusion or flow model directly on a native 3D latent, which may be a vector set~\citep{shape2vecset,michelangelo,clay,triposg,hunyuan3d2,hunyuan3d21,step1x3d} or a triplane~\citep{direct3d}, or a sparse or structured voxel grid~\citep{trellis,trellis2,sparc3d,hi3dgen}. All 4 non-agentic methods we evaluate build on the structured latent of TRELLIS~\citep{trellis}. A separate line generates low-polygon, artist-style meshes autoregressively, one face or one token at a time~\citep{polygen,meshgpt,meshxl,meshanything,meshanythingv2,pivotmesh,llamamesh,edgerunner,bpt,deepmesh,treemeshgpt,nautilus,meshsilksong,fastmesh,meshtron} or by flow matching in a mesh latent space~\citep{meshflow,polyflow}. These methods target compact mesh structure rather than instruction-guided changes to an existing asset. TAR3D~\citep{tar3d} autoregressively predicts quantized triplane tokens, while SeamFlow~\citep{seamflow} predicts cuts for UV unwrapping on an existing mesh. Part-aware modeling includes segmentation~\citep{sampart3d,partfield,p3sam}, generative decomposition and completion~\citep{holopart,bang,xpart,sculpt}, and generation of objects as sets of parts~\citep{partgen,partcrafter,omnipart,partpacker,autopartgen,kaininja,cubepart,megaparts}. Assembler~\citep{assembler} places input parts into a complete object, while SIMART and PWM-ArtGen address articulated assets~\citep{simart,pwmartgen}. Texture synthesis is a line of work of its own~\citep{texture,text2tex,paint3d,texturedreamer,mvpainter}, including PBR material generation~\citep{assetgen,materialanything,materialmvp,meshgen,pacture,matlat,texgaussian}. EditHero applies retexturing to one part while requiring the rest of the object to remain unchanged. CMD~\citep{cmd} supports progressive generation and local editing by conditioning on known parts. Its progressive examples illustrate construction, whereas EditHero supplies a benchmark of different instructions and exact reference states along each chain. EditHero uses the parts of an object both as the unit of editing and to build the exact ground truth.

\paragraph{Additional 3D editing methods.}
The methods we evaluate come from the 2 main families built on native 3D latents. Trained methods learn from paired edits. Much as ControlNet~\citep{controlnet} conditions an image model, they feed the source latent to the generator through an added control branch. PartFlow~\citep{partflow} does this with a source-control branch, and 3DEditFormer~\citep{editformer} does it with dual-guidance attention~\citep[see also][]{cai2025native,gat2026shapeup,yin2026editverse3d}. Training-free methods reuse pretrained generators without edit-specific training. Inversion-based methods such as VoxHammer~\citep{voxhammer} invert the source along the sampling trajectory and reuse its cached latents and attention keys and values in the preserved region. Inversion-free methods follow the difference between the source-conditioned and target-conditioned flows. Nano3D~\citep{nano3d} does this with the FlowEdit formulation~\citep{kulikov2025flowedit}, and related variants follow the same idea~\citep{zhou2026anchorflow,nie2026editflow3d,easy3e}. Single-view regeneration from the target render, which we report as a reference, is the limiting case of regenerating the object from an edited image. Further approaches are discussed below.
Early editing methods modify a neural field or a Gaussian splat with a 2D diffusion prior~\citep{haque2023instructnerf2nerf,sella2023voxe,zhuang2023dreameditor,chen2024gaussianeditor,wang2024gaussianeditor,khalid2024latenteditor,zhuang2024tipeditor,wu2024gaussctrl,chen2024dge}, and some of them take sketches~\citep{mikaeili2023sked,sketchdream} or drag handles~\citep{qu2025dragyourgaussian} as conditions instead of text. Recent methods also use mesh reconstruction or native 3D latents~\citep{chen2024shapeditor,chen2024mvedit,qi2024tailor3d,kim2025meshup,barda2025instant3dit,parelli2026threedlatte,zhou2026anchorflow,gat2026shapeup,hsiao2026vecsetedit,chi2026vinedresser3d,chen2026omni3dedit,yin2026editverse3d,ye2026hunyuan3dbuffalo,dreamedit3d,meshregen}. Other systems edit Gaussian scenes or scene atlases~\citep{gaussedit,chatedit3dpp}. Object insertion in 3D scenes and 3D object inpainting are also studied separately~\citep{freeinsert,objfiller3d}. Part-level and interactive methods let a user act on one part at a time~\citep{dong2024interactive3d,dong2024coin3d,wang2025partxmllm,sella2026proxe,slim2026compose}, and sketch-based systems have supported iterative refinement since Teddy~\citep{teddy,fibermesh,sketch2cad,free2cad,sketch2mesh,sens,gasketching,doodleyour3d,meshpad,pasta,sketchfacegs,scribblesense}. Sketch2CAD~\citep{sketch2cad} and MeshPad~\citep{meshpad} support sequential sketch-based modeling. Their evaluations focus on sketch interpretation and interactive modeling, whereas EditHero evaluates a sequence of natural-language part edits against exact 3D references.

\paragraph{2D image editing and its evaluation.}
Instruction-based image editing developed from inversion, personalization and attention control~\citep{sdedit,blendeddiffusion,prompt2prompt,textualinversion,dreambooth,nulltext,diffedit,imagic,pnp,masactrl,controlnet,leditspp} to models trained on edit pairs~\citep{instructpix2pix,emuedit,instructdiffusion,mgie,ultraedit,hqedit,anyedit,omniedit,picobanana} and to unified generators that edit in context~\citep{omnigen,icedit,step1xedit,fluxkontext,qwenimage,bagel}. Geometry-aware image editing uses 3D cues for spatial transformations or object removal~\citep{thinkinginboxes,geoedit,georemover}, and BlenderFusion~\citep{blenderfusion} combines 3D scene manipulation with generative image compositing. Before diffusion, iterative editing was studied as human-in-the-loop optimization, in which Bayesian optimization searches the parameters of an image or an animation under human preference feedback~\citep{brochu2010,prefbo,koyama2017sls,sequentialgallery}. Recent work applies the same idea to generative image search or adapter merging~\citep{hitlbo,gimmbo}. These systems study iterative improvement under human feedback. Benchmarks for image editing~\citep{editbench,magicbrush,editval,i2ebench,imgedit,risebench,krisbench,seeddataedit} established that ``did it edit'' and ``did it leave the rest alone'' must be scored separately, and that similarity to the source mixes up the requested change with drift. Multi-turn image evaluations use reference images or a VLM judge~\citep{magicbrush,multiturn2d,ma2026extensive,mteditflow,seqdecomp,omniiiebench,edival}. DiffSeg30k~\citep{diffseg30k} instead uses multi-turn edits to benchmark localized detection of generated content. MagicBrush provides intermediate target images reviewed by human annotators~\citep{magicbrush}. EditHero instead supplies complete 3D reference states that can be rebuilt deterministically, with the edited parts known at every turn. Multi-turn methods respond with memory of earlier turns or reinforcement learning over the sequence~\citep{nonmarkov,anchoredit,editr2}. Repeated image replication can accumulate artifacts that no-reference quality metrics fail to capture~\citep{banana100}. Related work studies different sources of accumulated error: repeated training on synthetic outputs~\citep{alemohammad2024mad,curationbackfires} and exposure bias within diffusion sampling~\citep{ning2024exposure}. We study successive 3D edits at inference time, where holes, loose pieces and displacement can be computed directly.

\paragraph{Additional LLM/VLM and agentic methods.}
LLMs can build 3D content by writing procedural code or calling tools. They do this for scenes and layouts~\citep{threedgpt,llmr,scenecraft,holodeck,layoutgpt,scenemotifcoder} and for editing an existing scene in Blender from renders~\citep{blenderalchemy,lu2025ll3m,wang2026ezblender,visculpt}. Related code-based systems generate objects and CAD programs~\citep{threedpremise,shapelib,meshcoder,text2cad,cadrecode,cadcodeverify,cadassistant,articulateanything}. Recent systems treat the program as the asset itself, with an agent that plans, writes and revises procedural code under visual feedback~\citep{procedura,adsl,nova3d,simworlds,vibeworlding,zerotocad}. EmbodiedGen V2~\citep{embodiedgen2} describes this as vibe coding of 3D worlds through a dialogue of instructions, aiming to make local edits while preserving the rest of the state. EditHero measures this workflow. Agentic pipelines exist for images as well~\citep{genartist,magicquill,t2icopilot,comfybench,talk2image,agentbanana,imageeditr1,iterativeeditagent,retouchiq,ape,agenticvisgen}. Related VLM-guided editing models and unified generators use learned multimodal reasoning~\citep{smartedit,got,uniworld}. Benchmarks for these agents~\citep{gu2025blendergym,gao2026threedcodebench} score the final state against one goal.

\paragraph{Ground truth by construction.}
Part-annotated collections~\citep{shapenet,partnet,partnext,fullpart} and procedural asset generators~\citep{articraft,infinigen} provide objects and parts for assembly. EditHero builds its chains with a program, like synthetic benchmarks whose targets are produced by a program~\citep{clevr,kubric,procthor} and unlike annotated datasets. The assembly engine that produces each state is deterministic, and rebuilding a turn from its logged operations reproduces the state exactly. The engine does not guarantee quality, so every chain went through manual review (Appendix~\ref{app:construction}).

\paragraph{Comparison with existing benchmarks.}
3DEditVerse~\citep{editformer} combines pose changes with image-guided appearance edits, while Nano3D-Edit-100k~\citep{nano3d} uses a native 3D editing method guided by edited images. Uni3DEdit-Bench~\citep{partflow} uses semantic-part transformations, and Delta3D~\citep{preintner2026threedmorph} removes parts from CAD assemblies. Steer3D\textquotesingle s EDIT3D-BENCH~\citep{ma2025steer3d} and the editing subset of 3D-Alpaca~\citep{ye2025shapellmomni} reconstruct edited images into target assets. Edit3D-Bench~\citep{voxhammer}, Eval3DEdit~\citep{zhou2026anchorflow}, BenchUp~\citep{gat2026shapeup} and EditFlow-Bench~\citep{nie2026editflow3d} have no target asset, so they score region preservation and alignment to an edited image instead. Scene-level benchmarks follow the same pattern~\citep{patro2026editbench3d,liu2025edit3r,zhu2026jointedit3d}. Each sample specifies a single editing goal, represented by an asset or an edited image. BlenderGym~\citep{gu2025blendergym} and 3DCodeBench~\citep{gao2026threedcodebench} let an agent iterate toward one fixed goal, and EZBlender~\citep{wang2026ezblender} bundles several edits into one prompt. Table~\ref{tab:bench-compare} places EditHero next to these benchmarks. The table compares target construction, instruction sequences, intermediate references and evaluation on each method's own outputs.

\begin{table}[ht]
\centering\footnotesize\setlength{\tabcolsep}{3.5pt}
\caption{EditHero compared with existing editing benchmarks. \emph{Target}: how the reference state is obtained (\emph{2D lift}: an edited image reconstructed as a 3D target; \emph{native edit}: a native 3D editing method guided by an edited image; \emph{none}: no target asset, scored against an edited image). \emph{Turns}: \emph{1}, one instruction per sample; \emph{goal}, one target task, the agent may iterate; \emph{bundle}, several edits given in one prompt; \emph{sequence}, different instructions applied in order. \emph{Ref.}: a reference is supplied after each instruction in a multi-turn sequence; a dash also covers single-turn or single-goal settings where this property does not apply. \emph{Own}: turn $k$ is applied to the method's output of turn $k-1$.}
\label{tab:bench-compare}
\begin{tabularx}{\linewidth}{@{}>{\raggedright\arraybackslash}X>{\raggedright\arraybackslash}p{1.5cm}>{\raggedright\arraybackslash}p{2.2cm}>{\raggedright\arraybackslash}p{1.6cm}cc@{}}
\toprule
Benchmark & Domain & Target & Turns & Ref. & Own \\
\midrule
MagicBrush~\citep{magicbrush} & 2D image & annotated image & sequence (few) & \checkmark & \checkmark \\
3DEditVerse~\citep{editformer} & 3D & pose edits / 2D lift & 1 & -- & -- \\
Nano3D-Edit-100k~\citep{nano3d} & 3D & native edit & 1 & -- & -- \\
Uni3DEdit-Bench~\citep{partflow} & 3D & semantic-part edits & 1 & -- & -- \\
EDIT3D-BENCH (Steer3D)~\citep{ma2025steer3d} & 3D & 2D lift & 1 & -- & -- \\
Delta3D~\citep{preintner2026threedmorph} & 3D & part removal & 1 & -- & -- \\
3D-Alpaca~\citep{ye2025shapellmomni} & 3D & 2D lift & 1 & -- & -- \\
Edit3D-Bench~\citep{voxhammer}, Eval3DEdit~\citep{zhou2026anchorflow} & 3D & none & 1 & -- & -- \\
BenchUp~\citep{gat2026shapeup}, EditFlow-Bench~\citep{nie2026editflow3d} & 3D & none & 1 & -- & -- \\
BlenderGym~\citep{gu2025blendergym}, 3DCodeBench~\citep{gao2026threedcodebench} & 3D scene, code & goal state & goal & -- & \checkmark \\
EZBlender~\citep{wang2026ezblender} & 3D scene & -- & bundle & -- & -- \\
\midrule
EditHero (ours) & 3D, part-level & program-built state & sequence (3--30) & \checkmark & \checkmark \\
\bottomrule
\end{tabularx}
\end{table}

\section{Benchmark construction details}
\label{app:construction}
\paragraph{Why assemble the targets?}
Assembling a target needs a suitable part and a plausible placement. The part has to make sense on the object (for example, a hat on a head or a barrel on a porch) and be attached where a person would attach it. Accepted placements must also satisfy explicit geometric constraints and stay reproducible after a repair. We therefore combine deterministic placement checks with manual review of every chain.

\paragraph{Hosts and slots.}
Our part library mainly contains PartVerse-XL~\citep{fullpart}, whose objects come from Objaverse-XL~\citep{objaversexl}. We also use very few parts from HY3D-Bench~\citep{hy3dbench}. They come without texture, so we texture them with the same pipeline as the retexture turns. For retrieval, we generate part descriptions with Qwen3-VL-32B-Instruct~\citep{qwen3vl}. These descriptions are separate from the editing instructions that people later review. Each chain starts from a textured, part-segmented object, which we call the \emph{host}. Its parts are grouped into a small number of \emph{slots}. A slot is a sub-assembly, such as a head, an arm or a porch canopy, that an instruction can name and that is large enough to see in a render. Slots are proposed automatically from the part contact graph and the part captions, and then accepted, merged or renamed by a reviewer looking at highlighted renders. Reviewers also correct the proposed slot names so they describe the grouped parts clearly.

\paragraph{Operations.}
\emph{Add} places a new part from the library on a free face of an existing part, \emph{remove} deletes the part or part group occupying a named slot, \emph{replace} swaps the part in a slot for a retrieved one on the same contact face, and \emph{retexture} repaints a slot and leaves its geometry exactly as it was. The new texture comes from a render of the slot restyled by Qwen-Image-Edit~\citep{qwenimage} and is applied to the unchanged slot geometry by the texturing module of TRELLIS.2~\citep{trellis2}. Before a support part is replaced, its attachments are removed in separate turns, and they are reattached in separate turns as well. Each step thus stays an explicit operation on a named part or part group. Instructions use natural-language part descriptions and never contain internal slot identifiers or library metadata.

\paragraph{Placement.}
Placement is deterministic, as in procedural datasets whose labels come from the generating program~\citep{clevr,kubric,infinigen}. For a replacement or an add-on part, the engine retrieves candidates by the similarity of their caption embeddings to the requested part~\citep{minilm,sbert}. It computes an initial pose from the host's contact face (a planar, axial or point mate, depending on the slot), scales the part so that its mount matches the face (between 0.75 and 2 times its original size) and snaps it into contact. Automatic candidate screening then applies 4 checks. A contact frame must exist, the remaining gap must be below 0.01 of the object diagonal, the object must stay one connected piece under that tolerance, and the interpenetration fraction must stay below 0.02. The automatic search rejects candidates that fail a check and tries the next one. Human review then assesses whether the assembly is visually appropriate, including intentional overlaps or disconnected components in stylized hosts.

\paragraph{Chain recipe.}
Each chain is stored as a JSON file that names the host and has one entry per turn. An entry gives the slot, the library part for an addition or a replacement, the accepted pose and, for a retexture, the generated texture. Slots that a turn does not name stay as they were. With the part library, the file rebuilds any state $S_k=o_k(\cdots o_1(S_0))$ on its own, without repeating retrieval or texture generation. A repair edits the entry of one turn, and that turn and the later ones are rebuilt and checked again.

\paragraph{Human review and repair.}
Every turn was rendered from several views before and after the edit and inspected. Reviewers flagged wrong part semantics (a foot placed as a hand), wrong orientation or mounting, implausible scale, instructions that gave away the source or were ambiguous, and dependency errors (an attachment silently carried along when its parent was replaced). Reviewers corrected poses in the operation log, saved revised geometry or materials as separate assets when needed, and rebuilt the affected turns. Original source assets were retained. Turns that could not be fixed were cut, and the rest of the chain was rebuilt from that point.

\paragraph{Families.}
Chains are grouped by the operations they contain. Family \emph{A} holds mostly additions. Families \emph{R} and \emph{P} contain only removals and replacements, respectively. Family \emph{M} contains only material edits. Family \emph{G} mixes geometry operations, and family \emph{X} mixes geometry and material operations. The benchmark contains 141 A, 72 M, 45 R, 54 P, 85 G and 60 X chains. Taking the 6 families together, we try to keep the share of each operation stable and even from turn to turn. Figure~\ref{fig:stats}(c) counts turns 1 to 9 one by one and puts the few edits from turn 10 onward into one bar.

\section{Evaluation protocol details}
\label{app:protocol}

\paragraph{Self-rollout.}
We give each method the initial mesh and, at turn $k$, the instruction together with the conditions it needs, such as a target image or a mask. Its input at turn $k$ is its own output at turn $k-1$. The main evaluation never resets the state to the ground truth, and Appendix~\ref{app:gtreset} reports a separate diagnostic that does. If a method fails a turn, for example with an empty or unloadable output, the rest of that chain is blocked and the chain counts as a failure. This rule is the same for the non-agentic methods and the LLM/VLM agents, and we never restart from the ground truth to rescue a chain. Infrastructure faults are handled separately (Appendix~\ref{app:setup}).

\paragraph{Conditioning.}
Every method receives the same target image at a given turn, rendered with the chain's fixed conditioning camera, and VoxHammer also receives an edit mask. The target image shows what the object should look like after the edit.

\paragraph{A shared coordinate frame and fixed views.}
We measure all states of a chain, ground truth and outputs alike, in one frame normalized by the bounding box of all its ground-truth states. We render them with one camera rig fitted to the bounding sphere of those states. The rig has 4 views 90$^\circ$ apart at one elevation, a fixed focal length and fixed lighting. If we re-centered every turn or every part, a part that moved would look as if it had stayed in place, and a new camera at every turn would make image metrics incomparable across turns. One default direction for all chains, however, can hide the edited part, since a small addition on the far side of the object barely shows in the conditioning view. We therefore choose the direction per chain. For every turn, the edited region is the set of part nodes whose geometry or material differs between the previous and the current ground-truth state. We fix for the chain the direction that makes its worst turn most visible (Appendix~\ref{app:camera}). To estimate visibility, we cast rays from surface samples of that region toward each candidate direction against the full state. Because the rig is fitted to all states of the chain, its framing reflects how large the object will become. No method receives the rig as geometry, and every method still sees the object in its own frame, fixed at the initial state.

\paragraph{Why these definitions.}
The definitions follow the design of Section~\ref{sec:protocol}, and Figure~\ref{fig:metric-cases} shows examples of the problems they avoid.

\emph{Score the edit and the rest separately.} A turn changes only a small part of an object. On the whole object, returning the initial object can therefore score higher than doing the requested edit on a state that has already drifted (Figure~\ref{fig:metric-cases}b). We therefore split every metric into $\editR{k}$, $\keepR{k}$ and the whole object. We take the regions from the named parts rather than from voxel or pixel differences, because a retexture leaves the ground-truth geometry unchanged even when a method changes it (Figure~\ref{fig:metric-cases}d). Using the named parts in both the old and the new state covers all 4 operations with one definition. The one-cell dilation absorbs a part that lands one cell off, so a small misplacement lowers IF but is not counted a second time as damage in CC.

\emph{Measure each turn against the method's own previous output.} Suppose a chain removes a hat at turn 2 and adds it back at turn 5. If we took the required addition from the ground truth, that is, the voxels present at turn 5 and absent at turn 4, the no-op output would already contain the hat and get full credit, although it never did anything. We instead take the required addition from the method's own previous output. The no-op output then needs nothing and gets no credit, and a method that did remove the hat gets credit only when it puts the hat back. For the same reason, a part lost at an earlier turn does not count as a successful removal now (Figure~\ref{fig:metric-cases}a). IF also catches a replacement that only deletes the old part. Its IF is close to 0 because the new part is missing, while its F-score to the target is within 0.02 of the no-op baseline (Figure~\ref{fig:metric-cases}c). CC$^{\text{prev}}$ uses the same anchor and asks whether this turn disturbed what it should keep. CC$^0$ compares with the initial object and shows how far the untouched content has drifted over the whole chain.

\emph{Read every score next to the references.} The no-op baseline has IF $=0$ and CC $=1$, and the regeneration baseline has a high IF and a low CC. A good editor should get close to both at once. We still report the target similarity $m_{\editR{k}}(\predS{k},\gtS{k})$, because prior work uses it. Read next to IF, it separates looking like the target from actually making the change at this turn.

\begin{figure}[ht]
\centering
\ehfit{\input{sec/fig_metric_cases}}
\caption{\textbf{Why the metrics are defined on regions and on the model's own state.} Four turns from the evaluation, in the conditioning view. Outlines mark the edited parts before and after the turn, dashed on the model's own states. For each turn, the left card shows how a target-similarity or ground-truth-based score can mislead, and the right card shows the corresponding IF or CC score (Eqs.~(\ref{eq:regions3d})--(\ref{eq:cc})). The tag on the right names the design choice the turn motivates.}
\label{fig:metric-cases}
\end{figure}

\section{Metric implementation}
\label{app:metrics}
\paragraph{3D operators.}
For object states $X$ (output) and $Y$ (reference), $\vox{\cdot}$ is the surface voxelization into $\grid$, $\pts{\cdot}$ is a dense set of surface samples, and $d(p,Y)=\min_{q\in\pts{Y}}\lVert p-q\rVert$ is the distance from a point to the reference surface (searched over all of $Y$, not only inside $R$). A sample belongs to $R$ when its voxel does.
\begin{equation}
\begin{aligned}
\IoUon{R}{X}{Y}
  &= \frac{\lvert\vox{X}\cap\vox{Y}\cap R\rvert}
           {\lvert(\vox{X}\cup\vox{Y})\cap R\rvert},\\[6pt]
\Recallon{R}{X}{Y}
  &= \frac{\lvert\vox{X}\cap\vox{Y}\cap R\rvert}
           {\lvert\vox{Y}\cap R\rvert}.
\end{aligned}
\end{equation}
\begin{equation}
\CDon{R}{X}{Y}
  = \frac12\left(
      \avg{p\in\pts{X}\cap R} d(p,Y)
      + \avg{q\in\pts{Y}\cap R} d(q,X)
    \right).
\end{equation}
\begin{equation}
\begin{aligned}
\Fon{R}{X}{Y}
  &= \frac{2\,\mathrm{Pr}\,\mathrm{Re}}{\mathrm{Pr}+\mathrm{Re}},\\[6pt]
\mathrm{Pr}
  &= \avg{p\in\pts{X}\cap R}\mathbb{I}\bigl[d(p,Y)<\tau\bigr],\\[4pt]
\mathrm{Re}
  &= \avg{q\in\pts{Y}\cap R}\mathbb{I}\bigl[d(q,X)<\tau\bigr].
\end{aligned}
\end{equation}

\paragraph{2D operators.}
For renders $I$ (output) and $J$ (reference) in the same fixed view, with RGB values in $[0,1]$,
\begin{equation}
\begin{aligned}
\MAEon{R}{I}{J}
  &= \avg{p\in R}\frac13\lVert I(p)-J(p)\rVert_1,\\[6pt]
\PSNRon{R}{I}{J}
  &= -10\log_{10}\left(
       \avg{p\in R}\frac13\lVert I(p)-J(p)\rVert_2^2
     \right),
\end{aligned}
\end{equation}
and $\SSIMon{R}{I}{J}$, $\LPIPSon{R}{I}{J}$ are the SSIM~\citep{ssim} and LPIPS~\citep{lpips} maps of $(I,J)$ averaged over the pixels of $R$. $\DINOon{R}{I}{J}$ is the cosine similarity of the DINOv2 patch features of $I$ and $J$, averaged over the patches whose center lies in $R$.

\paragraph{Constants.}
Voxelization uses the $64^3$ grid of the chain frame. Chamfer distance and F-score use 20{,}000 surface samples per mesh, and a sample belongs to a region when its voxel does. Nearest neighbors are searched over the whole reference, distances are in units of the chain side, and the F-score threshold is $\tau=0.01$. IF$^{+}$ (IF$^{-}$) is defined only when $A_k$ ($D_k$) has at least 8 voxels and at least a quarter of the ground-truth addition (removal) $\vox{\gtS{k}}\setminus\vox{\gtS{k-1}}$ ($\vox{\gtS{k-1}}\setminus\vox{\gtS{k}}$). Otherwise that turn has no IF. Defined-turn counts for the 2 main comparisons are in Table~\ref{tab:if-counts}. Image metrics use the 4 fixed views, composited on black, with the LPIPS map of AlexNet~\citep{lpips}, the SSIM map~\citep{ssim} and the patch features of DINOv2 ViT-B/14~\citep{oquab2023dinov2} (a $16\times16$ patch grid), each averaged over the region. Target-similarity metrics compare with the ground truth, and image drift compares consecutive outputs.

\section{Camera search}
\label{app:camera}
For each chain and each candidate camera direction, we sample points on the edited region of every turn and cast rays toward the camera against the full state. The visible projected area, weighted by the cosine to the view direction and corrected for depth, is divided by the frame area at the object center. The chosen direction is the one whose worst turn is most visible, with a small preference for the default direction. The estimate correlates with the measured pixel change of the rendered views at Spearman $\rho = 0.95$.

\section{Experimental setup and coverage}
\label{app:setup}
We evaluate PartFlow~\citep{partflow}, Nano3D~\citep{nano3d}, 3DEditFormer~\citep{editformer}, and VoxHammer~\citep{voxhammer} with their released code and weights. Each method is wrapped in an adapter that only supplies inputs and records outputs. The sampling code, seeds and hyperparameters are the authors' own. PartFlow, Nano3D and 3DEditFormer take the current state and the target render. VoxHammer also takes the ground-truth edit mask. Nano3D's official interface has add, remove and replace modes but no material mode. With the official routing, Nano3D rejects a material instruction and the rest of the chain is blocked, which would happen on all \nMaterialChains{} chains that contain one. We therefore route every material instruction to the replace mode. The adapter passes the unchanged instruction text and the target render exactly as for a replace turn, and the output is scored like any other turn. The Nano3D rows use this routing on those \nMaterialChains{} chains and the official routing on the other chains. All methods are scored with the same fixed-camera protocol of Section~\ref{sec:protocol}. The meshes of the 4 methods, exported through TRELLIS, carry no metallic value and would render as dark metal, so we render them as non-metallic. The agents' outputs of Section~\ref{sec:agentic} set their own materials and are rendered as exported.

The TRELLIS regeneration baseline runs the released TRELLIS-image-large pipeline~\citep{trellis} on the target render of every turn, in the same fixed view that the non-agentic methods receive. It sees neither the instruction nor its previous output, so its turns are independent. Its output lies in the normalized frame of TRELLIS. One view does not fix the orientation in that frame, so on some turns the object comes out turned by a multiple of $90^\circ$. We therefore align every turn to the ground-truth state of the same turn with a yaw search followed by a similarity ICP. The ground-truth geometry is used only for this alignment. It completes all \nEvalEdits{} turns of the \nEvalChains{} chains.

Table~\ref{tab:coverage} reports, per method, the chains completed, the completed turns, the failed or subsequently blocked turns, and the failure categories. Infrastructure faults, such as a time-out or a lost node, were handled by rerunning the whole chain from its first turn, and the decision to rerun never depended on the outputs. These faults do not appear in the table. The remaining failures occurred during model execution. VoxHammer produces an empty sparse structure on 39 chains, and Nano3D writes no edited voxels on 2 chains. Nano3D would also fail on every material instruction without this adaptation, but with the replace routing above it completes all chains except those 2. We keep these as failures rather than drop the chains, because a method that cannot run a turn has not completed the chain.

\begin{table}[ht]
\centering\footnotesize
\caption{Execution coverage on the \nEvalChains{} chains. Completion means the chain ran to the end, not that the edits were correct. Conditions: \emph{image} = current state and target render; \emph{+ mask} = also the ground-truth edit mask. The Turns column reports completed turns out of all scheduled turns. Failed / blocked includes the failure turn and later turns that could not run.}
\label{tab:coverage}
\ehsep{9pt}\ehfit{\setlength{\tabcolsep}{3.5pt}
\begin{tabular}{@{}lllrr>{\raggedright\arraybackslash}p{2.75cm}@{}}
\toprule
Method & Cond. & Chains & Turns & Failed / blocked & Failure category \\
\midrule
PartFlow & image & 457/457 & 2755/2755 & 0 & -- \\
Nano3D & image & 455/457 & 2749/2755 & 6 & no edited voxels (2) \\
3DEditFormer & image & 457/457 & 2755/2755 & 0 & -- \\
VoxHammer & image + mask & 418/457 & 2569/2755 & 186 & empty sparse output (39) \\
\bottomrule
\end{tabular}
}
\end{table}

\begin{table}[ht]
\centering\footnotesize
\caption{Defined-turn counts for IF, by operation and comparison set. Retexture turns have no geometry IF and are excluded. These counts differ from execution coverage.}
\label{tab:if-counts}
\ehsep{14pt}\ehfitplain{\begin{tabular}{lrrrr}
\toprule
Method & All defined & Add & Remove & Replace \\
\midrule
\multicolumn{5}{l}{\emph{Main evaluation (457 chains)}} \\
PartFlow & 1878 & 960 & 414 & 504 \\
Nano3D & 1929 & 962 & 438 & 529 \\
3DEditFormer & 1896 & 966 & 414 & 516 \\
VoxHammer & 1754 & 894 & 376 & 484 \\
No-op baseline & 1815 & 946 & 386 & 483 \\
TRELLIS regen. & 1793 & 967 & 350 & 476 \\
\midrule
\multicolumn{5}{l}{\emph{Agentic/non-agentic comparison (55 chains)}} \\
PartFlow & 270 & 125 & 55 & 90 \\
Nano3D & 274 & 125 & 57 & 92 \\
3DEditFormer & 279 & 125 & 56 & 98 \\
VoxHammer & 249 & 122 & 50 & 77 \\
No-op baseline & 243 & 119 & 46 & 78 \\
TRELLIS regen. & 258 & 125 & 51 & 82 \\
GLM 5.3 Flash & 236 & 125 & 42 & 69 \\
DeepSeek V4.1 Flash & 268 & 125 & 54 & 89 \\
Sol 6 (medium) & 267 & 125 & 56 & 86 \\
Astra (medium) & 288 & 125 & 64 & 99 \\
Fable 5.1 & 283 & 125 & 59 & 99 \\
Opus 5.5 & 299 & 125 & 70 & 104 \\
\bottomrule
\end{tabular}
}
\end{table}

\section{Whole-object metrics, per-turn curves and end state}
\label{app:longhorizon}
Table~\ref{tab:main} reports whole-object and preservation metrics in 2 groups. The first group uses the chains that all 4 non-agentic methods complete, so every method is scored on the same turns. The second uses all completed turns of the methods that complete almost every chain. The metric means in the 2 groups agree to within 0.015. The no-op baseline has a higher surface IoU to the target than every method (0.71 against at most 0.58 on these chains) and a lower LPIPS to the target render. Nano3D has the highest IoU and preservation of the 4, and its IF is close to VoxHammer's. VoxHammer gets its IF while keeping only 0.45 of the untouched initial object (CC$^{0}$). The TRELLIS regeneration baseline, which builds every turn from the target render alone, sits at the other extreme. It has the highest IF on these chains (0.51) but also the lowest IoU (0.29) and the lowest preservation (CC$^{\text{prev}}$ 0.38).

\begin{table}[ht]
\centering\footnotesize
\caption{Whole-object metrics of the self-rollout runs. Upper block: the \cohortChains{} chains that all 4 non-agentic methods complete, so all rows share the same turns. Lower block: all completed turns of the \nEvalChains{} chains, for the methods with full or near-full coverage. Bold: best non-agentic method in the upper block.}
\label{tab:main}
\ehsep{10.5pt}\ehfit{\setlength{\tabcolsep}{4pt}
\begin{tabular}{@{}lrrrrrrr@{}}
\toprule
Method & $n$ & IoU$\uparrow$ & IF$\uparrow$ & CC$^{\mathrm{prev}}\uparrow$ & CC$^{0}\uparrow$ & LPIPS$\downarrow$ & drift$\downarrow$ \\
\midrule
\multicolumn{8}{@{}l}{\emph{416 chains completed by every non-agentic method}} \\
PartFlow & 2472 & 0.516 & 0.126 & 0.862 & 0.674 & 0.177 & 0.014 \\
Nano3D & 2472 & \textbf{0.584} & 0.318 & \textbf{0.887} & \textbf{0.729} & \textbf{0.160} & \textbf{0.009} \\
3DEditFormer & 2472 & 0.414 & 0.202 & 0.774 & 0.490 & 0.198 & 0.018 \\
VoxHammer & 2472 & 0.380 & \textbf{0.322} & 0.714 & 0.453 & 0.220 & 0.022 \\
No-op baseline & 2472 & 0.712 & 0.000 & 1.000 & 1.000 & 0.144 & 0.000 \\
TRELLIS regen. & 2472 & 0.294 & 0.506 & 0.384 & 0.280 & 0.203 & 0.041 \\
\midrule
\multicolumn{8}{@{}l}{\emph{All completed turns (methods with full or near-full coverage)}} \\
PartFlow & 2755 & 0.507 & 0.122 & 0.862 & 0.666 & 0.181 & 0.014 \\
Nano3D & 2749 & 0.575 & 0.313 & 0.887 & 0.718 & 0.163 & 0.009 \\
3DEditFormer & 2755 & 0.408 & 0.197 & 0.772 & 0.481 & 0.203 & 0.018 \\
No-op baseline & 2755 & 0.700 & 0.000 & 1.000 & 1.000 & 0.151 & 0.000 \\
\bottomrule
\end{tabular}
}
\end{table}

\begin{figure}[ht]
\centering
\ehfitplain{\begin{tikzpicture}
\begin{groupplot}[group style={group size=3 by 1,horizontal sep=0.9cm},width=0.36\linewidth,height=4.3cm,
  xmin=0.7,xmax=7.3,ymin=0,ymax=1,xtick={1,...,7},xlabel={turn},xlabel near ticks,
  tick label style={font=\scriptsize},label style={font=\scriptsize},title style={font=\scriptsize,yshift=-1ex},
  grid=major,grid style={gray!25},legend style={font=\tiny,draw=none,fill=none,at={(0.02,0.98)},anchor=north west,row sep=-2pt},
  every axis plot/.append style={line width=0.9pt,mark size=1.2pt}]
\nextgroupplot[title={Surface IoU to target}]
\addplot[name path=lo0c1,draw=none,forget plot] table[x=turn,y=lo]{figures/curves/iou_gt_PartFlow.dat};
\addplot[name path=hi0c1,draw=none,forget plot] table[x=turn,y=hi]{figures/curves/iou_gt_PartFlow.dat};
\addplot[c1,opacity=0.12,forget plot] fill between[of=lo0c1 and hi0c1];
\addplot[c1,mark=*] table[x=turn,y=mean]{figures/curves/iou_gt_PartFlow.dat};
\addplot[name path=lo0c2,draw=none,forget plot] table[x=turn,y=lo]{figures/curves/iou_gt_Nano3D.dat};
\addplot[name path=hi0c2,draw=none,forget plot] table[x=turn,y=hi]{figures/curves/iou_gt_Nano3D.dat};
\addplot[c2,opacity=0.12,forget plot] fill between[of=lo0c2 and hi0c2];
\addplot[c2,mark=*] table[x=turn,y=mean]{figures/curves/iou_gt_Nano3D.dat};
\addplot[name path=lo0c3,draw=none,forget plot] table[x=turn,y=lo]{figures/curves/iou_gt_3DEditFormer.dat};
\addplot[name path=hi0c3,draw=none,forget plot] table[x=turn,y=hi]{figures/curves/iou_gt_3DEditFormer.dat};
\addplot[c3,opacity=0.12,forget plot] fill between[of=lo0c3 and hi0c3];
\addplot[c3,mark=*] table[x=turn,y=mean]{figures/curves/iou_gt_3DEditFormer.dat};
\addplot[name path=lo0c4,draw=none,forget plot] table[x=turn,y=lo]{figures/curves/iou_gt_VoxHammer.dat};
\addplot[name path=hi0c4,draw=none,forget plot] table[x=turn,y=hi]{figures/curves/iou_gt_VoxHammer.dat};
\addplot[c4,opacity=0.12,forget plot] fill between[of=lo0c4 and hi0c4];
\addplot[c4,mark=*] table[x=turn,y=mean]{figures/curves/iou_gt_VoxHammer.dat};
\addplot[black,dashed,mark=none] table[x=turn,y=mean]{figures/curves/iou_gt_no-op.dat};
\nextgroupplot[title={Instruction following (IF)},legend to name=curveslegend,legend columns=6]
\addplot[name path=lo1c1,draw=none,forget plot] table[x=turn,y=lo]{figures/curves/if_self_PartFlow.dat};
\addplot[name path=hi1c1,draw=none,forget plot] table[x=turn,y=hi]{figures/curves/if_self_PartFlow.dat};
\addplot[c1,opacity=0.12,forget plot] fill between[of=lo1c1 and hi1c1];
\addplot[c1,mark=*] table[x=turn,y=mean]{figures/curves/if_self_PartFlow.dat};
\addlegendentry{PartFlow}
\addplot[name path=lo1c2,draw=none,forget plot] table[x=turn,y=lo]{figures/curves/if_self_Nano3D.dat};
\addplot[name path=hi1c2,draw=none,forget plot] table[x=turn,y=hi]{figures/curves/if_self_Nano3D.dat};
\addplot[c2,opacity=0.12,forget plot] fill between[of=lo1c2 and hi1c2];
\addplot[c2,mark=*] table[x=turn,y=mean]{figures/curves/if_self_Nano3D.dat};
\addlegendentry{Nano3D}
\addplot[name path=lo1c3,draw=none,forget plot] table[x=turn,y=lo]{figures/curves/if_self_3DEditFormer.dat};
\addplot[name path=hi1c3,draw=none,forget plot] table[x=turn,y=hi]{figures/curves/if_self_3DEditFormer.dat};
\addplot[c3,opacity=0.12,forget plot] fill between[of=lo1c3 and hi1c3];
\addplot[c3,mark=*] table[x=turn,y=mean]{figures/curves/if_self_3DEditFormer.dat};
\addlegendentry{3DEditFormer}
\addplot[name path=lo1c4,draw=none,forget plot] table[x=turn,y=lo]{figures/curves/if_self_VoxHammer.dat};
\addplot[name path=hi1c4,draw=none,forget plot] table[x=turn,y=hi]{figures/curves/if_self_VoxHammer.dat};
\addplot[c4,opacity=0.12,forget plot] fill between[of=lo1c4 and hi1c4];
\addplot[c4,mark=*] table[x=turn,y=mean]{figures/curves/if_self_VoxHammer.dat};
\addlegendentry{VoxHammer}
\addplot[black,dashed,mark=none] table[x=turn,y=mean]{figures/curves/if_self_no-op.dat};
\addlegendentry{no-op}
\nextgroupplot[title={Preservation of the initial object (CC$^{0}$)}]
\addplot[name path=lo2c1,draw=none,forget plot] table[x=turn,y=lo]{figures/curves/cc_turn0_dil_PartFlow.dat};
\addplot[name path=hi2c1,draw=none,forget plot] table[x=turn,y=hi]{figures/curves/cc_turn0_dil_PartFlow.dat};
\addplot[c1,opacity=0.12,forget plot] fill between[of=lo2c1 and hi2c1];
\addplot[c1,mark=*] table[x=turn,y=mean]{figures/curves/cc_turn0_dil_PartFlow.dat};
\addplot[name path=lo2c2,draw=none,forget plot] table[x=turn,y=lo]{figures/curves/cc_turn0_dil_Nano3D.dat};
\addplot[name path=hi2c2,draw=none,forget plot] table[x=turn,y=hi]{figures/curves/cc_turn0_dil_Nano3D.dat};
\addplot[c2,opacity=0.12,forget plot] fill between[of=lo2c2 and hi2c2];
\addplot[c2,mark=*] table[x=turn,y=mean]{figures/curves/cc_turn0_dil_Nano3D.dat};
\addplot[name path=lo2c3,draw=none,forget plot] table[x=turn,y=lo]{figures/curves/cc_turn0_dil_3DEditFormer.dat};
\addplot[name path=hi2c3,draw=none,forget plot] table[x=turn,y=hi]{figures/curves/cc_turn0_dil_3DEditFormer.dat};
\addplot[c3,opacity=0.12,forget plot] fill between[of=lo2c3 and hi2c3];
\addplot[c3,mark=*] table[x=turn,y=mean]{figures/curves/cc_turn0_dil_3DEditFormer.dat};
\addplot[name path=lo2c4,draw=none,forget plot] table[x=turn,y=lo]{figures/curves/cc_turn0_dil_VoxHammer.dat};
\addplot[name path=hi2c4,draw=none,forget plot] table[x=turn,y=hi]{figures/curves/cc_turn0_dil_VoxHammer.dat};
\addplot[c4,opacity=0.12,forget plot] fill between[of=lo2c4 and hi2c4];
\addplot[c4,mark=*] table[x=turn,y=mean]{figures/curves/cc_turn0_dil_VoxHammer.dat};
\addplot[black,dashed,mark=none] table[x=turn,y=mean]{figures/curves/cc_turn0_dil_no-op.dat};
\end{groupplot}
\node at ($(group c2r1.south)+(0,-0.95cm)$) {\pgfplotslegendfromname{curveslegend}};
\end{tikzpicture}}
\caption{Per-turn metrics on the \cohortChains{} chains that all 4 non-agentic methods complete (\cohortHosts{} hosts). Shaded: 95\% host-clustered bootstrap intervals. The dashed line is the no-op baseline, which has IF $=0$ and CC$^{0}=1$ by definition. Its IoU compares the unchanged initial object with the current target.}
\label{fig:curves}
\end{figure}
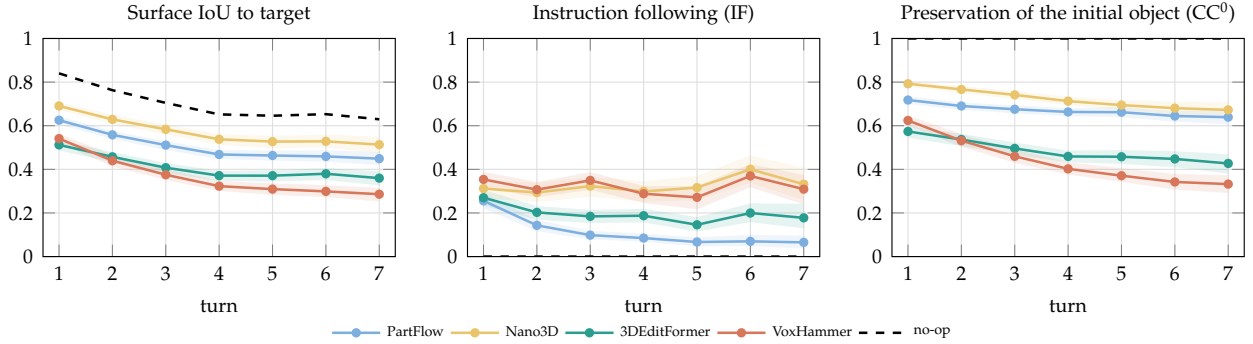

Figure~\ref{fig:curves} and Table~\ref{tab:perturn} follow the metrics turn by turn on the \cohortChains{} chains completed by all 4 non-agentic methods, with 95\% intervals from a bootstrap~\citep{efron1979} over hosts (chains built on the same host are not independent). Target similarity drops as the chain goes on. The no-op baseline also declines as the targets move away from the initial object, but it stays above all 4 methods. The TRELLIS regeneration baseline, whose turns are independent, stays almost flat (IoU 0.30 at turn~1 and 0.27 at turn~7). These curves alone cannot tell whether later targets are harder or the errors are inherited, and Appendix~\ref{app:gtreset} tests this. After turn 7 the set shrinks to 80 chains at turn~8 and we do not draw conclusions from it.

\begin{table}[ht]
\centering\footnotesize
\caption{Per-turn means within the matched cohort (turns 1--7), using chains that reach each turn and excluding undefined scores. TRELLIS regen.: the regeneration baseline, whose turns are independent.}
\label{tab:perturn}
\ehsep{11.5pt}\ehfit{\setlength{\tabcolsep}{4pt}
\begin{tabular}{@{}lrrrrrrr@{}}
\toprule
Turn & 1 & 2 & 3 & 4 & 5 & 6 & 7 \\
$n$ chains & 416 & 416 & 416 & 375 & 302 & 233 & 135 \\
\midrule
\multicolumn{8}{@{}l}{\itshape IoU} \\
\quad PartFlow & 0.625 & 0.558 & 0.511 & 0.468 & 0.463 & 0.459 & 0.449 \\
\quad Nano3D & 0.691 & 0.629 & 0.583 & 0.538 & 0.527 & 0.528 & 0.513 \\
\quad 3DEditFormer & 0.512 & 0.457 & 0.408 & 0.371 & 0.371 & 0.380 & 0.360 \\
\quad VoxHammer & 0.541 & 0.440 & 0.375 & 0.323 & 0.309 & 0.299 & 0.286 \\
\quad No-op baseline & 0.840 & 0.763 & 0.705 & 0.652 & 0.646 & 0.653 & 0.630 \\
\quad TRELLIS regen. & 0.303 & 0.298 & 0.297 & 0.293 & 0.290 & 0.283 & 0.274 \\
\addlinespace[2pt]
\multicolumn{8}{@{}l}{\itshape IF} \\
\quad PartFlow & 0.256 & 0.143 & 0.098 & 0.085 & 0.067 & 0.070 & 0.065 \\
\quad Nano3D & 0.312 & 0.294 & 0.323 & 0.299 & 0.316 & 0.400 & 0.332 \\
\quad 3DEditFormer & 0.270 & 0.203 & 0.184 & 0.188 & 0.146 & 0.200 & 0.178 \\
\quad VoxHammer & 0.354 & 0.307 & 0.349 & 0.288 & 0.272 & 0.370 & 0.309 \\
\quad No-op baseline & 0.000 & 0.000 & 0.000 & 0.000 & 0.000 & 0.000 & 0.000 \\
\quad TRELLIS regen. & 0.532 & 0.474 & 0.518 & 0.491 & 0.489 & 0.525 & 0.520 \\
\addlinespace[2pt]
\multicolumn{8}{@{}l}{\itshape CC$^{0}$} \\
\quad PartFlow & 0.718 & 0.690 & 0.675 & 0.663 & 0.662 & 0.645 & 0.639 \\
\quad Nano3D & 0.793 & 0.766 & 0.741 & 0.713 & 0.694 & 0.680 & 0.673 \\
\quad 3DEditFormer & 0.574 & 0.537 & 0.496 & 0.459 & 0.458 & 0.448 & 0.427 \\
\quad VoxHammer & 0.624 & 0.532 & 0.460 & 0.402 & 0.371 & 0.342 & 0.332 \\
\quad No-op baseline & 1.000 & 1.000 & 1.000 & 1.000 & 1.000 & 1.000 & 1.000 \\
\quad TRELLIS regen. & 0.300 & 0.292 & 0.289 & 0.275 & 0.273 & 0.254 & 0.247 \\
\addlinespace[2pt]
\bottomrule
\end{tabular}
}
\end{table}

\paragraph{End state.}
The per-turn curves stop at turn 7 and average over chains of different lengths, so they do not show where a chain ends up. Table~\ref{tab:endstate} scores the last turn of every chain of this set, grouped by chain length, next to the first turn of the same chains. All 4 methods lose similarity and preservation from the first to the last turn in every length group, and they lose more on the longest chains. The IoU of PartFlow falls by 0.10 on chains of 3 turns and by 0.20 on chains of 11 turns or more. The CC$^{0}$ of 3DEditFormer falls by 0.06 and 0.41 on the same groups, and that of VoxHammer by 0.11 and 0.38. The IoU of the no-op baseline falls with length too, from 0.83--0.88 at the first turn to 0.61--0.78 at the end, because the target moves away from the initial object. End-state IoU should be read against this baseline, and none of the 4 non-agentic methods beats it on mean end-state IoU in any length group. The TRELLIS regeneration baseline, which never carries a state over, has an IoU of roughly 0.3 at both the first and last turns.

\begin{table}[ht]
\centering\footnotesize
\caption{End-state evaluation on the same chains, by chain length: IoU to the target and CC$^{0}$ at the first turn $\to$ at the last turn of the chain (means over the chains in the group; the first and last groups are small).}
\label{tab:endstate}
\ehfit{\setlength{\tabcolsep}{3pt}
\begin{tabular}{@{}lrrrrrrrr@{}}
\toprule
Chain length & \multicolumn{2}{c}{3 turns (41 chains)} & \multicolumn{2}{c}{4--6 turns (240 chains)} & \multicolumn{2}{c}{7--10 turns (124 chains)} & \multicolumn{2}{c}{11--30 turns (11 chains)} \\
\cmidrule(lr){2-3} \cmidrule(lr){4-5} \cmidrule(lr){6-7} \cmidrule(lr){8-9}
Method & IoU & CC$^{0}$ & IoU & CC$^{0}$ & IoU & CC$^{0}$ & IoU & CC$^{0}$ \\
\midrule
PartFlow & 0.67\,$\to$\,0.57 & 0.71\,$\to$\,0.69 & 0.61\,$\to$\,0.43 & 0.72\,$\to$\,0.63 & 0.63\,$\to$\,0.43 & 0.72\,$\to$\,0.62 & 0.67\,$\to$\,0.47 & 0.72\,$\to$\,0.58 \\
Nano3D & 0.74\,$\to$\,0.68 & 0.79\,$\to$\,0.76 & 0.68\,$\to$\,0.50 & 0.80\,$\to$\,0.68 & 0.69\,$\to$\,0.50 & 0.78\,$\to$\,0.64 & 0.75\,$\to$\,0.50 & 0.81\,$\to$\,0.58 \\
3DEditFormer & 0.55\,$\to$\,0.46 & 0.59\,$\to$\,0.53 & 0.49\,$\to$\,0.33 & 0.55\,$\to$\,0.40 & 0.54\,$\to$\,0.36 & 0.60\,$\to$\,0.42 & 0.58\,$\to$\,0.24 & 0.65\,$\to$\,0.24 \\
VoxHammer & 0.58\,$\to$\,0.47 & 0.62\,$\to$\,0.51 & 0.54\,$\to$\,0.28 & 0.63\,$\to$\,0.35 & 0.54\,$\to$\,0.28 & 0.61\,$\to$\,0.33 & 0.56\,$\to$\,0.23 & 0.61\,$\to$\,0.23 \\
\midrule
No-op baseline & 0.88\,$\to$\,0.78 & 1.00\,$\to$\,1.00 & 0.83\,$\to$\,0.61 & 1.00\,$\to$\,1.00 & 0.85\,$\to$\,0.62 & 1.00\,$\to$\,1.00 & 0.86\,$\to$\,0.64 & 1.00\,$\to$\,1.00 \\
TRELLIS regen. & 0.29\,$\to$\,0.31 & 0.28\,$\to$\,0.30 & 0.30\,$\to$\,0.30 & 0.30\,$\to$\,0.28 & 0.31\,$\to$\,0.28 & 0.31\,$\to$\,0.24 & 0.28\,$\to$\,0.27 & 0.27\,$\to$\,0.21 \\
\bottomrule
\end{tabular}
}
\end{table}

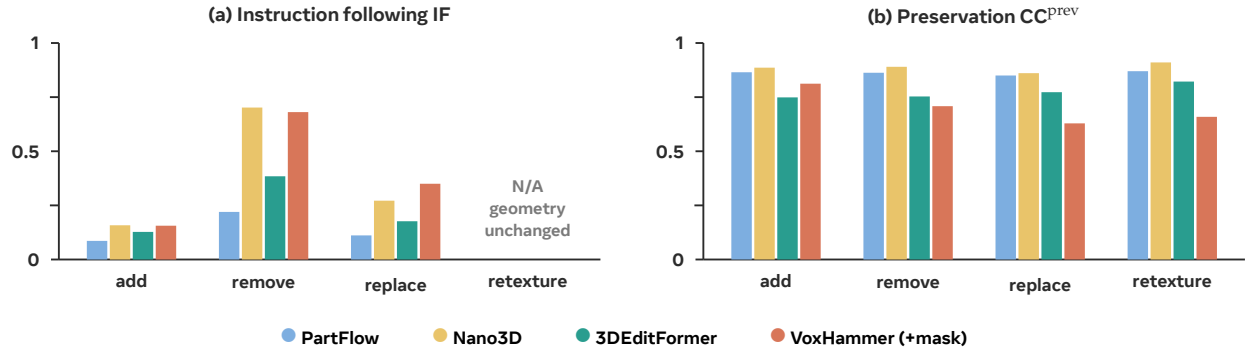
\begin{figure}[ht]
\centering
\setlength{\abovecaptionskip}{5pt}\setlength{\belowcaptionskip}{2pt}
\ehfit{\begin{tikzpicture}
\begin{groupplot}[eh-bar,group style={group size=2 by 1,horizontal sep=1.15cm},width=.46\linewidth,height=4cm,ymin=0,ymax=1,xmin=-.55,xmax=3.55,ytick={0,.5,1},yticklabels={0,0.5,1},minor ytick={.25,.75},xtick={0,1,2,3},xticklabels={add,remove,replace,retexture},xtick style={draw=none},ybar=0.8pt,/pgf/bar width=6.5pt]
\nextgroupplot[title={(a) Instruction following IF}]
\addplot[fill=c1,draw=none] coordinates {(0,0.0860) (1,0.2198) (2,0.1113)};
\addplot[fill=c2,draw=none] coordinates {(0,0.1581) (1,0.7019) (2,0.2717)};
\addplot[fill=c3,draw=none] coordinates {(0,0.1275) (1,0.3844) (2,0.1767)};
\addplot[fill=c4,draw=none] coordinates {(0,0.1561) (1,0.6808) (2,0.3499)};
\node[anchor=south,align=center,font=\sffamily\fontsize{6}{7}\selectfont,text=pl-mute] at (axis cs:3,0.04) {N/A\\geometry\\unchanged};
\nextgroupplot[title={(b) Preservation CC$^{\mathrm{prev}}$}]
\addplot[fill=c1,draw=none] coordinates {(0,0.8649) (1,0.8626) (2,0.8499) (3,0.8698)};
\addplot[fill=c2,draw=none] coordinates {(0,0.8861) (1,0.8901) (2,0.8607) (3,0.9103)};
\addplot[fill=c3,draw=none] coordinates {(0,0.7488) (1,0.7531) (2,0.7728) (3,0.8219)};
\addplot[fill=c4,draw=none] coordinates {(0,0.8121) (1,0.7082) (2,0.6288) (3,0.6590)};
\end{groupplot}
\path let \p1 = (current bounding box.south west), \p2 = (current bounding box.south east) in node[anchor=north, font=\sffamily\fontsize{6.4}{7.4}\selectfont, inner sep=0pt] at ({(\x1+\x2)/2}, {\y1 - 0.25cm}) {\textcolor{c1}{\scalebox{1.7}{$\bullet$}}\,PartFlow \hspace{14pt} \textcolor{c2}{\scalebox{1.7}{$\bullet$}}\,Nano3D \hspace{14pt} \textcolor{c3}{\scalebox{1.7}{$\bullet$}}\,3DEditFormer \hspace{14pt} \textcolor{c4}{\scalebox{1.7}{$\bullet$}}\,VoxHammer (+mask)};
\end{tikzpicture}}
\caption{\textbf{Metrics by operation.} Means over completed turns with defined scores. Retexture IF is undefined (N/A, not zero). Coverage and conditioning differ by method, so the methods are not compared on the same turns here.}
\label{fig:operations}
\end{figure}

\section{Own state versus ground-truth state}
\label{app:gtreset}
The decline under self-rollout has 2 possible causes. Later instructions may be harder, or errors in the previous output may make them harder to execute. To tell them apart we rerun Nano3D and 3DEditFormer on \gtResetChains{} chains and change one thing. The input state at turn $k$ is the ground-truth state $S_{k-1}$ instead of the model's own $\hat S_{k-1}$, with the same instruction, target render, camera and seed. The ground-truth states are encoded into each method's latent space with the same pipeline that encodes its initial object. Under this condition the turns are independent. Table~\ref{tab:gtreset} pairs the 2 input conditions on the same chains at each turn. Figure~\ref{fig:turns}(c) instead uses a fixed subset of \gtResetFixedChains{} chains with at least 5 turns, so its means differ slightly from those in the table.

In this diagnostic, much of the decline comes from the inherited state. Under self-rollout, IoU to the target falls from turn 1 to turn 5 (0.68 to 0.52 for Nano3D and 0.52 to 0.36 for 3DEditFormer). With the ground-truth input it stays at 0.68 for Nano3D and falls only to 0.46 for 3DEditFormer. Instruction following is higher with the ground-truth input at every turn after the first, by 0.06 to 0.08 for Nano3D and by 0.08 to 0.19 for 3DEditFormer (95\% host-clustered intervals exclude zero at turns 2 to 5). For this diagnostic, $\mathrm{CC}^{\mathrm{prev}}_{\Delta}$ measures preservation outside the one-cell-dilated voxel difference between the 2 ground-truth states, rather than the part-based region of Eq.~\eqref{eq:cc}. Preservation of the previous state moves the other way. Over turns 2 to 7, for Nano3D it is 0.90--0.94 under self-rollout and 0.71--0.75 with the ground-truth input, and for 3DEditFormer it is 0.80--0.83 against 0.46--0.56. In other words, a method keeps its own output well but disturbs a state it did not produce, as it also does at turn 1. A high CC$^{\text{prev}}$ under self-rollout only means that the method keeps its own state, which can already be wrong.

\begin{table}[ht]
\centering\footnotesize
\caption{Self-rollout versus ground-truth input, paired per turn. Turn 1 has the same input in both conditions. \emph{self-rollout}: the model consumes its own previous output; \emph{ground-truth input}: it consumes the ground-truth state. Turn $k$ averages the chains with at least $k$ turns (200 at turns 1--3, 181 at turn 4, 147 at turn 5, 115 at turn 6 and 65 at turn 7), the same chains in both conditions.}
\label{tab:gtreset}
\ehsep{9pt}\ehfit{\setlength{\tabcolsep}{4pt}
\begin{tabular}{@{}llrrrrrrr@{}}
\toprule
Metric & State & 1 & 2 & 3 & 4 & 5 & 6 & 7 \\
\midrule
\multicolumn{9}{@{}l}{\itshape Nano3D} \\
\quad IoU & self-rollout & 0.684 & 0.619 & 0.575 & 0.529 & 0.518 & 0.525 & 0.534 \\
 & ground-truth input & 0.684 & 0.694 & 0.676 & 0.666 & 0.681 & 0.694 & 0.725 \\
\quad IF & self-rollout & 0.335 & 0.320 & 0.306 & 0.291 & 0.277 & 0.365 & 0.387 \\
 & ground-truth input & 0.337 & 0.378 & 0.364 & 0.356 & 0.337 & 0.424 & 0.466 \\
\quad CC$^{\mathrm{prev}}_{\Delta}$ & self-rollout & 0.790 & 0.903 & 0.906 & 0.896 & 0.899 & 0.900 & 0.936 \\
 & ground-truth input & 0.789 & 0.750 & 0.730 & 0.714 & 0.727 & 0.739 & 0.751 \\
\addlinespace[2pt]
\multicolumn{9}{@{}l}{\itshape 3DEditFormer} \\
\quad IoU & self-rollout & 0.506 & 0.446 & 0.405 & 0.367 & 0.364 & 0.383 & 0.380 \\
 & ground-truth input & 0.506 & 0.482 & 0.471 & 0.431 & 0.460 & 0.502 & 0.545 \\
\quad IF & self-rollout & 0.282 & 0.197 & 0.201 & 0.172 & 0.152 & 0.184 & 0.185 \\
 & ground-truth input & 0.284 & 0.289 & 0.279 & 0.316 & 0.321 & 0.361 & 0.379 \\
\quad CC$^{\mathrm{prev}}_{\Delta}$ & self-rollout & 0.561 & 0.799 & 0.810 & 0.813 & 0.826 & 0.821 & 0.834 \\
 & ground-truth input & 0.560 & 0.515 & 0.502 & 0.456 & 0.486 & 0.531 & 0.564 \\
\bottomrule
\end{tabular}
}
\end{table}

\section{Material edits}
\label{app:material}
\begin{figure}[ht]
\centering
\ehfitplain{\input{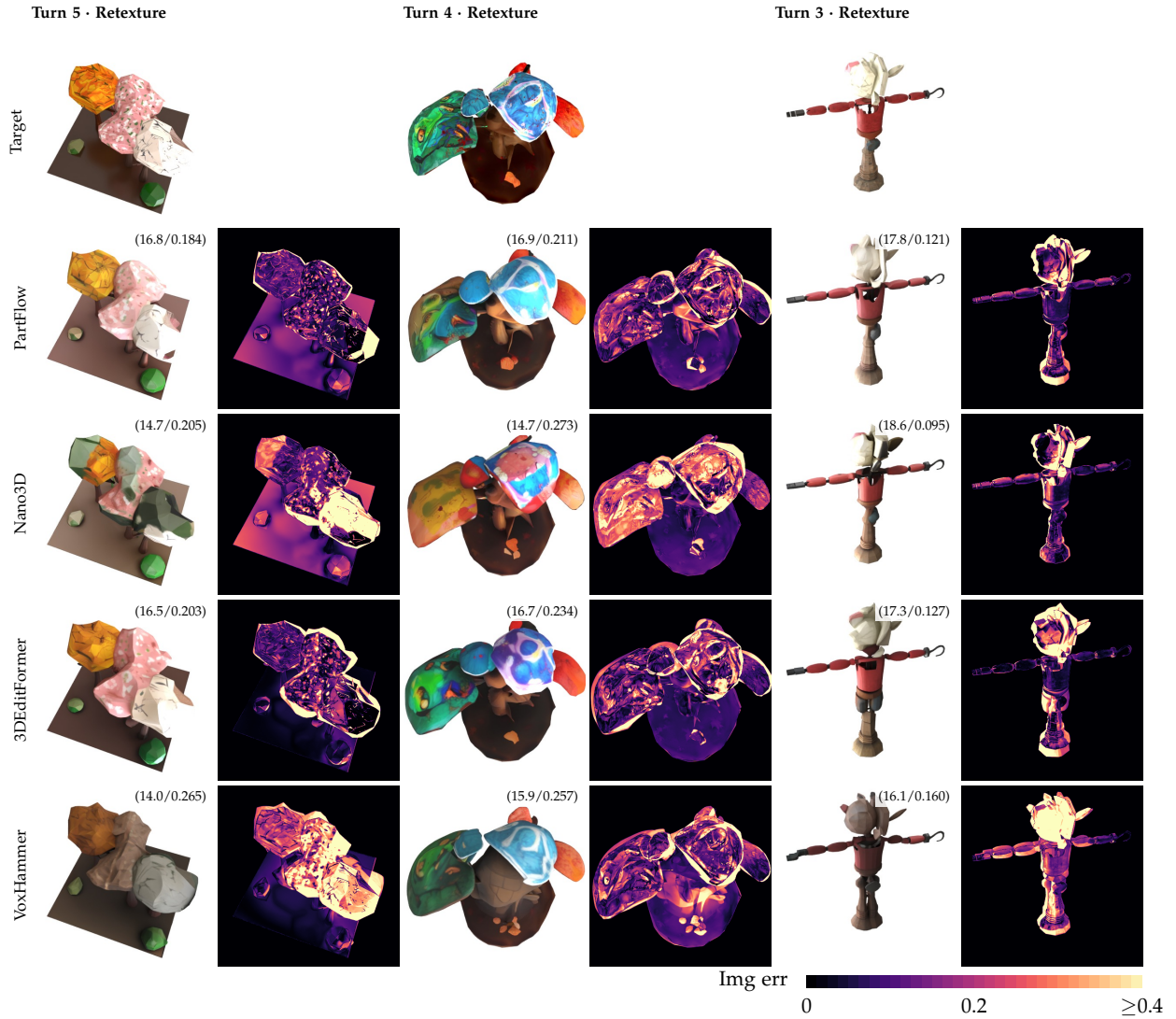}}
\caption{Retexture turns from 3 chains in the conditioning view. For each method the left tile is its render and the right tile its per-pixel error to the target render (mean absolute RGB difference; color bar at the bottom right). The numbers in the top-right corner are (PSNR in dB / LPIPS) of the render against the target view. The ground truth changes only the appearance of the named part, while the errors of the 4 methods spread over the whole object.}
\label{fig:material}
\end{figure}

Geometry metrics do not register material edits, so we measure them in images. For every retexture turn we take the pixels where the ground-truth render changes. This changed-pixel diagnostic uses a narrower region than the part-based definition in Section~\ref{sec:protocol}. We then measure how much each method's own consecutive renders change there, relative to the ground-truth change (Table~\ref{tab:material}). PartFlow, 3DEditFormer and VoxHammer produce 0.68--0.98 times the ground-truth image-change magnitude and bring the result closer to the target. Nano3D, whose material turns run in the replace mode, changes the region by \matrepOwnOverGT{} times the ground-truth image change, and its distance to the target is close to that of its own unedited previous output. It completes all chains containing retexture turns and has the highest whole-object IoU on those chains (0.67). This does not show material accuracy, since retexturing leaves the ground-truth geometry unchanged. Figure~\ref{fig:material} shows what the other methods do instead. They change the appearance of untouched parts too, so their errors spread over the whole object instead of staying on the named part. Their colors also shift away from the target.

\begin{table}[ht]
\centering\footnotesize
\caption{Change inside the ground-truth changed-pixel region on retexture turns. Views count turn $\times$ view pairs. A view is skipped when the ground-truth change covers fewer than 50 pixels, and VoxHammer has fewer views because some of its chains stop early. Own change / GT change compares a method's consecutive renders with the ground-truth change. Residuals are mean absolute error to the target render inside the region, for the method's render (\emph{method}) and for its previous output (\emph{no edit}).}
\label{tab:material}
\ehsep{12pt}\ehfit{\setlength{\tabcolsep}{4pt}
\begin{tabular}{@{}lrrrr@{}}
\toprule
 & & Own change & \multicolumn{2}{c}{Residual to target} \\
\cmidrule(lr){4-5}
Method & Views & / GT change & method & no edit \\
\midrule
PartFlow & 2563 & 0.68 & 0.187 & 0.213 \\
Nano3D & 2563 & 0.31 & 0.209 & 0.210 \\
3DEditFormer & 2563 & 0.80 & 0.186 & 0.219 \\
VoxHammer & 2384 & 0.98 & 0.216 & 0.225 \\
\bottomrule
\end{tabular}
}
\end{table}

\section{LLM/VLM setup and cost}
\label{app:llm}
\paragraph{Setup.}
Opus~5.5 and Fable~5.1 run in Claude Code, Astra and Sol~6 run as Codex subagents, and GLM~5.3 Flash and DeepSeek~V4.1 Flash run through OpenRouter in our API tool loop, which dispatches the tools a model requests and returns their results to the conversation. We evaluate each model with the tools provided by its runtime. Each chain is edited by its own agent, which starts a fresh session and shares no memory with other chains. Each agent receives the source mesh, the chain's instructions and target renders, and the camera parameters. It edits its own previous output and saves a full mesh after every turn. All models use the same base editing prompt (Appendix~\ref{app:prompts}) and our adaptation of img2threejs v2.0.0~\citep{img2threejs}, which retains staged construction and render-and-compare refinement while editing an existing mesh and using Blender for rendering. New parts are built in code, removals delete the named part, and retextures change its material while keeping its geometry. External assets, hidden ground-truth meshes, and generative 3D or image models are prohibited.

\paragraph{Budgets.}
Opus~5.5 and Fable~5.1 were asked to finish a chain in about 3 hours. GLM~5.3 Flash and DeepSeek~V4.1 Flash had 200 model responses and 3 hours per chain, followed by a completion pass without a cap. Astra and Sol~6 use medium reasoning effort without a time cap. We report the cost they actually used.

\paragraph{Coverage and failures.}
A turn counts as delivered when its mesh exists, is not empty and loads. Otherwise the turn fails and the rest of the chain is blocked, as for the non-agentic methods.

\paragraph{Cost.}
Table~\ref{tab:cost} reports the amortized cost per edit for all 6 LLMs. We divide each chain's recorded time, tokens and model calls by its number of delivered turns, then take the median across chains. Opus~5.5 uses \amortOpusSec{}~s per edit (90th percentile \amortOpusPninety{}~s) and \amortOpusCalls{} model calls, most of whose input is cached context.

\begin{table}[ht]
\centering\footnotesize
\caption{Cost per delivered edit. Non-agentic methods: median time per completed turn on one local H100 (90th percentile in parentheses). All agent costs are chain totals divided by delivered turns, summarized by the median across 55 chains (90th percentile for time). Input tokens include cached context. Agent time includes model and tool latency and is not comparable to local GPU time.}
\label{tab:cost}
\ehsep{12pt}\ehfitplain{\begin{tabular}{lrrr}
\toprule
Method & time per turn (s) & tokens per turn (in / out) & calls per turn \\
\midrule
PartFlow & \secPartflow{} & -- & -- \\
Nano3D & \secNano{} & -- & -- \\
3DEditFormer & \secEditformer{} & -- & -- \\
VoxHammer & \secVox{} & -- & -- \\
\midrule
GLM~5.3 Flash & \amortGLMSec{} (\amortGLMPninety{}) & \amortGLMTokIn{} / \amortGLMTokOut{} & \amortGLMCalls{} \\
DeepSeek~V4.1 Flash & \amortDeepSeekSec{} (\amortDeepSeekPninety{}) & \amortDeepSeekTokIn{} / \amortDeepSeekTokOut{} & \amortDeepSeekCalls{} \\
Sol~6 (medium) & \amortSolSec{} (\amortSolPninety{}) & \amortSolTokIn{} / \amortSolTokOut{} & \amortSolCalls{} \\
Astra (medium) & \amortAstraSec{} (\amortAstraPninety{}) & \amortAstraTokIn{} / \amortAstraTokOut{} & \amortAstraCalls{} \\
Fable~5.1 & \amortFableSec{} (\amortFablePninety{}) & \amortFableTokIn{} / \amortFableTokOut{} & \amortFableCalls{} \\
Opus~5.5 & \amortOpusSec{} (\amortOpusPninety{}) & \amortOpusTokIn{} / \amortOpusTokOut{} & \amortOpusCalls{} \\
\bottomrule
\end{tabular}}
\end{table}

\section{Extended metrics and held-out views}
\label{app:ext}
Tables~\ref{tab:ext} and~\ref{tab:ext-img} add metrics that the main tables leave out, computed on the same runs. For geometry we report the symmetric Chamfer distance~\citep{fan2017pointset} and the 95th percentile of the two-sided nearest-neighbor distance (HD95~\citep{huttenlocher1993,taha2015}), which is more sensitive to large surface errors than the mean distance. We also add normal consistency (NC~\citep{mescheder2019occ}) and the F-score at 2 thresholds~\citep{knapitsch2017,tatarchenko2019what}. The same HD95 and NC are also restricted to an untouched region (subscript u), which here lies outside the dilated voxel difference $\vox{\gtS{0}}\triangle\vox{\gtS{k}}$ between the initial and the target state. This region can include parts that were edited and later restored, whereas the cumulative part-based region of Eq.~\eqref{eq:cc} excludes them. The last geometry column ($\Delta$frag.) is the median difference in the number of small connected components (below 0.5\% of surface area) between the prediction and ground truth. Negative values mean fewer such components, which is not always a better reconstruction. For images we report PSNR, LPIPS~\citep{lpips}, silhouette IoU from the alpha channel and DINOv2 ViT-B/14 cosine similarity~\citep{oquab2023dinov2} to the target render. PSNR, LPIPS and silhouette IoU are given separately for the conditioning view (subscript 0), which every image-conditioned method sees, and for the 3 held-out views 90$^\circ$ apart (subscript 1-3), which are not provided as target images. DINO$_{\mathrm{retex}}$ is the similarity in the conditioning view on retexture turns only. All numbers are means over completed turns, except $\Delta$frag., which is a median.

\begin{figure}[ht]
\centering
\ehfit{\input{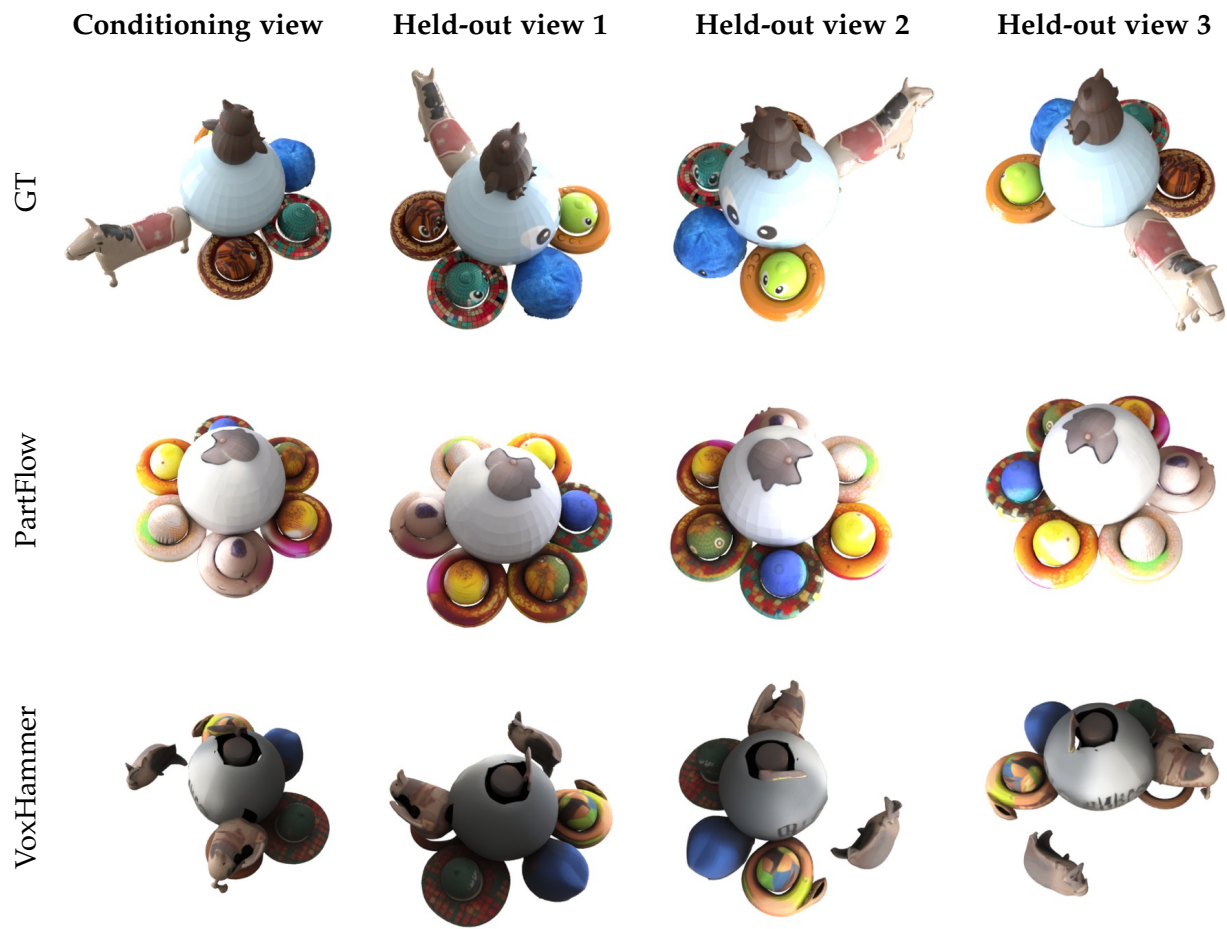}}
\caption{\textbf{Conditioning and held-out views.} GT, PartFlow and VoxHammer at turn 6 of one chain, in the conditioning view and the 3 held-out views.}
\label{fig:multiview}
\end{figure}

The 4 non-agentic methods show little average gap between conditioning and held-out views (PSNR within 0.5~dB, LPIPS within 0.01). In the untouched region the no-op baseline has an HD95 of 0.009, and the 4 methods have values 3 to 8 times as large. On retexture turns, DINO similarity is close to the no-op baseline (0.851) for PartFlow (0.854), Nano3D (0.843) and 3DEditFormer (0.847), and lower for VoxHammer (0.804).

\begin{table}[ht]
\centering\footnotesize\setlength{\tabcolsep}{4pt}
\caption{Extended geometry metrics over completed turns on the \nEvalChains{} chains. Subscript u: untouched region; $\Delta$frag.: median fragment-count difference relative to ground truth.}
\label{tab:ext}
\ehsep{8pt}\ehfit{\begin{tabular}{lrrrrrrrr}
\toprule
Method & CD$\downarrow$ & HD95$\downarrow$ & NC$\uparrow$ & F$_{.01}\uparrow$ & F$_{.02}\uparrow$ & HD95$_{\mathrm{u}}\downarrow$ & NC$_{\mathrm{u}}\uparrow$ & $\Delta$frag. \\
\midrule
PartFlow & 0.022 & 0.114 & 0.81 & 0.67 & 0.80 & 0.039 & 0.87 & -17.0 \\
Nano3D & 0.020 & 0.100 & 0.83 & 0.71 & 0.83 & 0.034 & 0.87 & -17.0 \\
3DEditFormer & 0.028 & 0.117 & 0.75 & 0.55 & 0.73 & 0.065 & 0.79 & -17.0 \\
VoxHammer & 0.027 & 0.122 & 0.75 & 0.52 & 0.69 & 0.072 & 0.79 & -17.0 \\
\midrule
No-op baseline & 0.020 & 0.122 & 0.85 & 0.77 & 0.84 & 0.009 & 0.93 & 0.0 \\
\bottomrule
\end{tabular}
}
\end{table}

\begin{table}[ht]
\centering\footnotesize\setlength{\tabcolsep}{2.6pt}
\caption{Multi-view image metrics over completed turns. Subscript 0: conditioning view; 1-3: held-out views; retex: retexture turns only.}
\label{tab:ext-img}
\ehsep{4.5pt}\ehfit{\begin{tabular}{lrrrrrrrr}
\toprule
Method & PSNR$_0$ & PSNR$_{1\text{-}3}$ & LPIPS$_0\downarrow$ & LPIPS$_{1\text{-}3}\downarrow$ & Sil$_0\uparrow$ & Sil$_{1\text{-}3}\uparrow$ & DINO$_{1\text{-}3}\uparrow$ & DINO$_{\mathrm{retex}}\uparrow$ \\
\midrule
PartFlow & 17.2 & 17.0 & 0.180 & 0.182 & 0.78 & 0.79 & 0.753 & 0.854 \\
Nano3D & 17.8 & 17.7 & 0.163 & 0.163 & 0.82 & 0.81 & 0.765 & 0.843 \\
3DEditFormer & 16.5 & 16.0 & 0.197 & 0.205 & 0.73 & 0.73 & 0.752 & 0.847 \\
VoxHammer & 15.3 & 15.2 & 0.221 & 0.219 & 0.72 & 0.72 & 0.738 & 0.804 \\
\midrule
No-op baseline & 18.1 & 18.9 & 0.161 & 0.147 & 0.79 & 0.80 & 0.752 & 0.851 \\
\bottomrule
\end{tabular}
}
\end{table}

\section{Conditioning and held-out views by region}
\label{app:views}
The main tables average every image metric over the 4 fixed views. Tables~\ref{tab:view-lpips} and~\ref{tab:view-psnr} split that average by view and by region on the comparison subset of Section~\ref{sec:agentic}, so the LLM/VLM agents appear next to the non-agentic methods. The conditioning view is the one in which each method receives the target render, and no method ever sees the 3 held-out views. Most methods score a little better in the conditioning view, and the gap is widest in the edit region. There Opus~5.5 reaches an LPIPS of 0.215 in the conditioning view and 0.246 in the held-out views, while the non-agentic methods lose at most 0.017. The no-op baseline goes the other way. The camera direction of each chain is chosen so that every edit is visible in the conditioning view, which makes an unedited object look worse there than from the side. The order of the methods is almost the same in both kinds of view, with only 2 neighboring pairs swapping places, so averaging the 4 views does not favor a method that fits the view it was shown.

\begin{table}[ht]
\centering\footnotesize\setlength{\tabcolsep}{4pt}
\caption{LPIPS by view and region on the comparison subset. Cond.: the conditioning view; held-out: mean of the 3 other fixed views. Means over each method's completed turns. Best method (excluding the 2 baselines) in bold.}
\label{tab:view-lpips}
\ehfit{\begin{tabular}{l cc cc cc}
\toprule
 & \multicolumn{2}{c}{Edit region} & \multicolumn{2}{c}{Unchanged region} & \multicolumn{2}{c}{Whole object} \\
\cmidrule(lr){2-3}\cmidrule(lr){4-5}\cmidrule(lr){6-7}
Method & Cond.$\downarrow$ & Held-out$\downarrow$ & Cond.$\downarrow$ & Held-out$\downarrow$ & Cond.$\downarrow$ & Held-out$\downarrow$ \\
\midrule
No-op baseline & 0.592 & 0.558 & 0.140 & 0.128 & 0.165 & 0.151 \\
TRELLIS regen. & 0.380 & 0.422 & 0.189 & 0.225 & 0.200 & 0.237 \\
\midrule
\multicolumn{7}{@{}l}{\itshape Non-agentic methods} \\
PartFlow & 0.497 & 0.503 & 0.161 & 0.169 & 0.178 & 0.186 \\
Nano3D & 0.438 & 0.452 & 0.138 & 0.145 & 0.156 & 0.163 \\
3DEditFormer & 0.462 & 0.475 & 0.162 & 0.180 & 0.177 & 0.195 \\
VoxHammer & 0.478 & 0.495 & 0.195 & 0.200 & 0.209 & 0.214 \\
\midrule
\multicolumn{7}{@{}l}{\itshape LLM/VLM agents} \\
GLM~5.3 Flash & 0.458 & 0.443 & 0.123 & 0.117 & 0.141 & 0.134 \\
DeepSeek~V4.1 Flash & 0.356 & 0.369 & 0.087 & 0.093 & 0.102 & 0.108 \\
Sol~6 (medium) & 0.332 & 0.346 & 0.076 & 0.080 & 0.090 & 0.094 \\
Astra (medium) & 0.272 & 0.292 & 0.061 & 0.066 & 0.073 & 0.079 \\
Fable~5.1 & 0.278 & 0.299 & 0.063 & 0.069 & 0.075 & 0.082 \\
Opus~5.5 & \textbf{0.215} & \textbf{0.246} & \textbf{0.048} & \textbf{0.056} & \textbf{0.058} & \textbf{0.067} \\
\bottomrule
\end{tabular}
}
\end{table}

\begin{table}[ht]
\centering\footnotesize\setlength{\tabcolsep}{4pt}
\caption{PSNR (dB) by view and region on the comparison subset, as in Table~\ref{tab:view-lpips}.}
\label{tab:view-psnr}
\ehfit{\begin{tabular}{l cc cc cc}
\toprule
 & \multicolumn{2}{c}{Edit region} & \multicolumn{2}{c}{Unchanged region} & \multicolumn{2}{c}{Whole object} \\
\cmidrule(lr){2-3}\cmidrule(lr){4-5}\cmidrule(lr){6-7}
Method & Cond.$\uparrow$ & Held-out$\uparrow$ & Cond.$\uparrow$ & Held-out$\uparrow$ & Cond.$\uparrow$ & Held-out$\uparrow$ \\
\midrule
No-op baseline & 10.6 & 10.4 & 23.0 & 23.8 & 17.5 & 18.4 \\
TRELLIS regen. & 13.6 & 12.0 & 16.5 & 15.3 & 15.8 & 14.8 \\
\midrule
\multicolumn{7}{@{}l}{\itshape Non-agentic methods} \\
PartFlow & 10.4 & 9.8 & 18.3 & 18.0 & 17.0 & 16.8 \\
Nano3D & 12.1 & 11.4 & 19.2 & 19.0 & 17.7 & 17.6 \\
3DEditFormer & 10.8 & 10.1 & 18.1 & 17.4 & 17.0 & 16.3 \\
VoxHammer & 11.0 & 10.3 & 16.5 & 16.4 & 15.7 & 15.6 \\
\midrule
\multicolumn{7}{@{}l}{\itshape LLM/VLM agents} \\
GLM~5.3 Flash & 16.9 & 16.4 & 24.5 & 25.2 & 20.8 & 21.4 \\
DeepSeek~V4.1 Flash & 20.2 & 19.3 & 27.2 & 27.6 & 23.9 & 24.1 \\
Sol~6 (medium) & 20.6 & 19.5 & 28.3 & 28.3 & 24.5 & 24.6 \\
Astra (medium) & 21.6 & 20.6 & 29.0 & 29.3 & 25.4 & 25.7 \\
Fable~5.1 & 21.6 & 20.4 & 29.0 & 28.9 & 25.3 & 25.4 \\
Opus~5.5 & \textbf{23.2} & \textbf{21.7} & \textbf{30.3} & \textbf{29.9} & \textbf{26.8} & \textbf{26.6} \\
\bottomrule
\end{tabular}
}
\end{table}

\clearpage
\section{More qualitative comparisons}
\label{app:qual-methods}
\begin{figure}[p]
\centering
\ehfitpage{\input{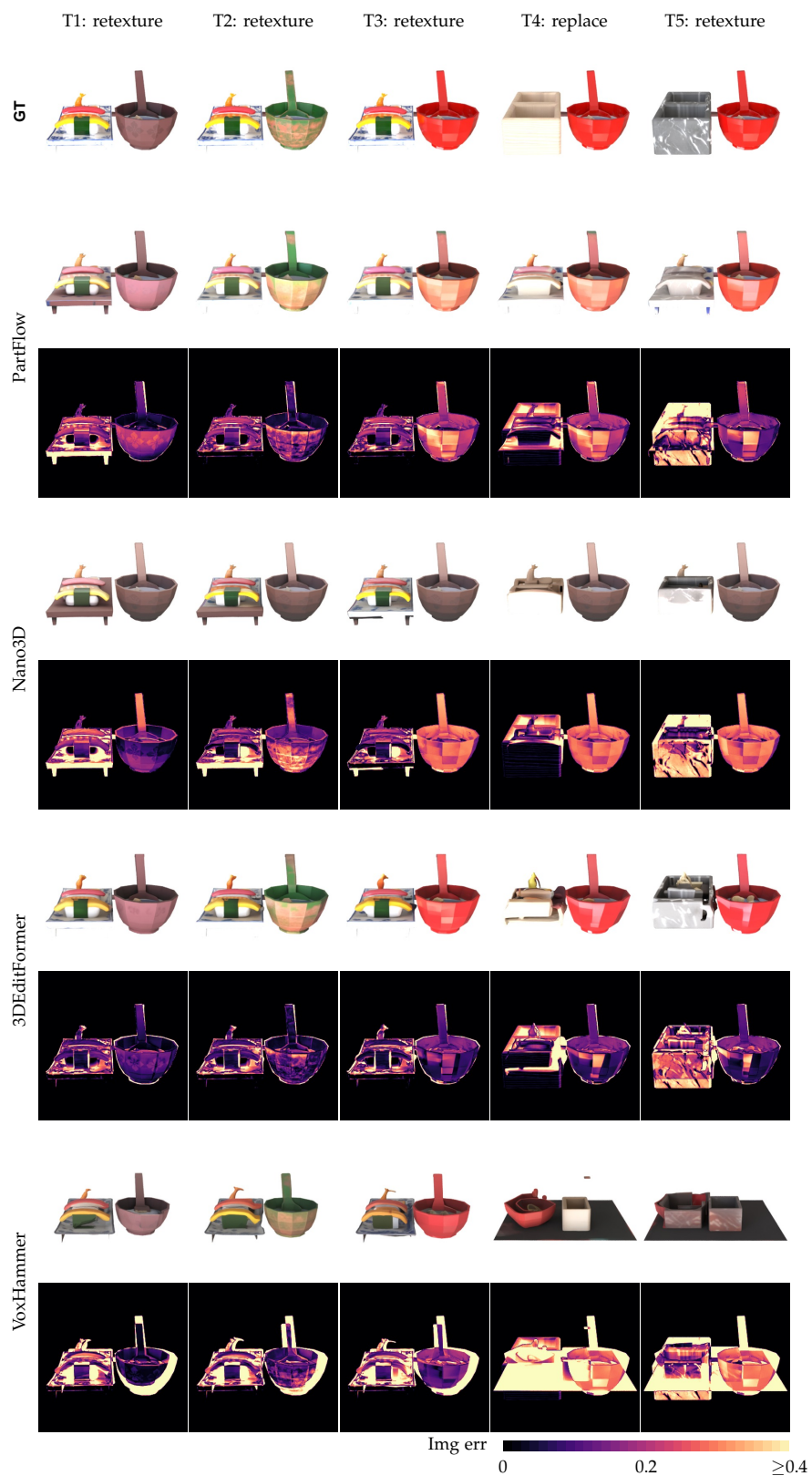}}
\caption{\textbf{A table setting} (turns 1--5): turns 1--3 and 5 change the material of the sushi tray, the soup bowl or the compartment tray, and turn 4 replaces the sushi tray and the sushi with a wooden compartment tray. Shown in the conditioning view.}
\label{fig:qual-heat}
\end{figure}

\begin{figure}[p]
\centering
\ehfitpage{\input{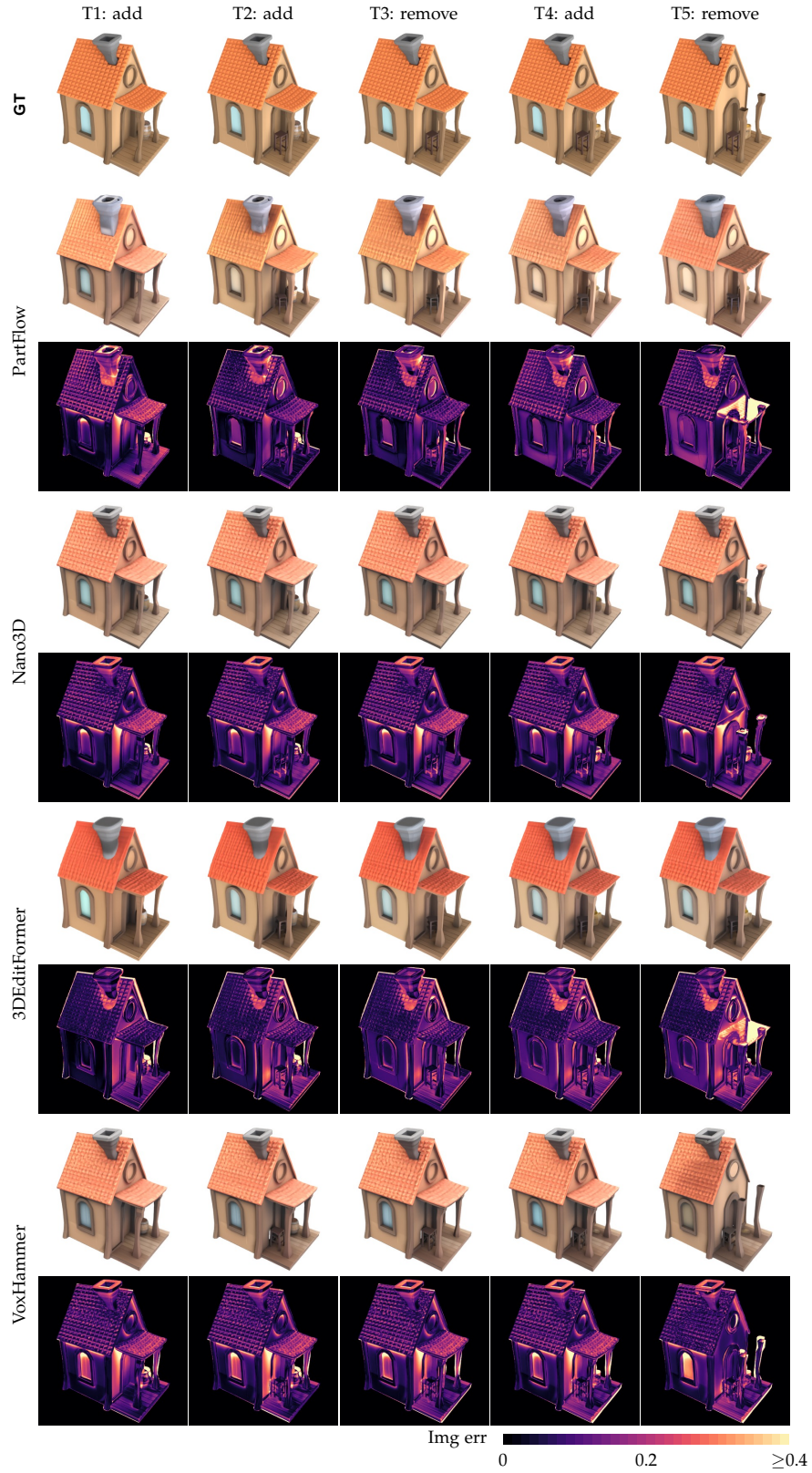}}
\caption{\textbf{A house} (turns 1--5): 3 additions and 2 removals on the porch. Shown from a held-out view other than the conditioning view.}
\label{fig:qual-house}
\end{figure}

Figures~\ref{fig:qual-heat} and~\ref{fig:qual-house} follow 2 more chains. One is a table setting whose edits are mostly material changes, and the other is a house whose porch gains and loses parts. Rows are methods and columns are the first 5 turns. Under each output is its per-pixel error to the ground truth of that turn (mean absolute RGB difference).

\section{Qualitative comparison of LLM/VLM agents and non-agentic methods}
\label{app:qual-llm}
\begin{figure}[p]
\centering
\ehfitpage{\input{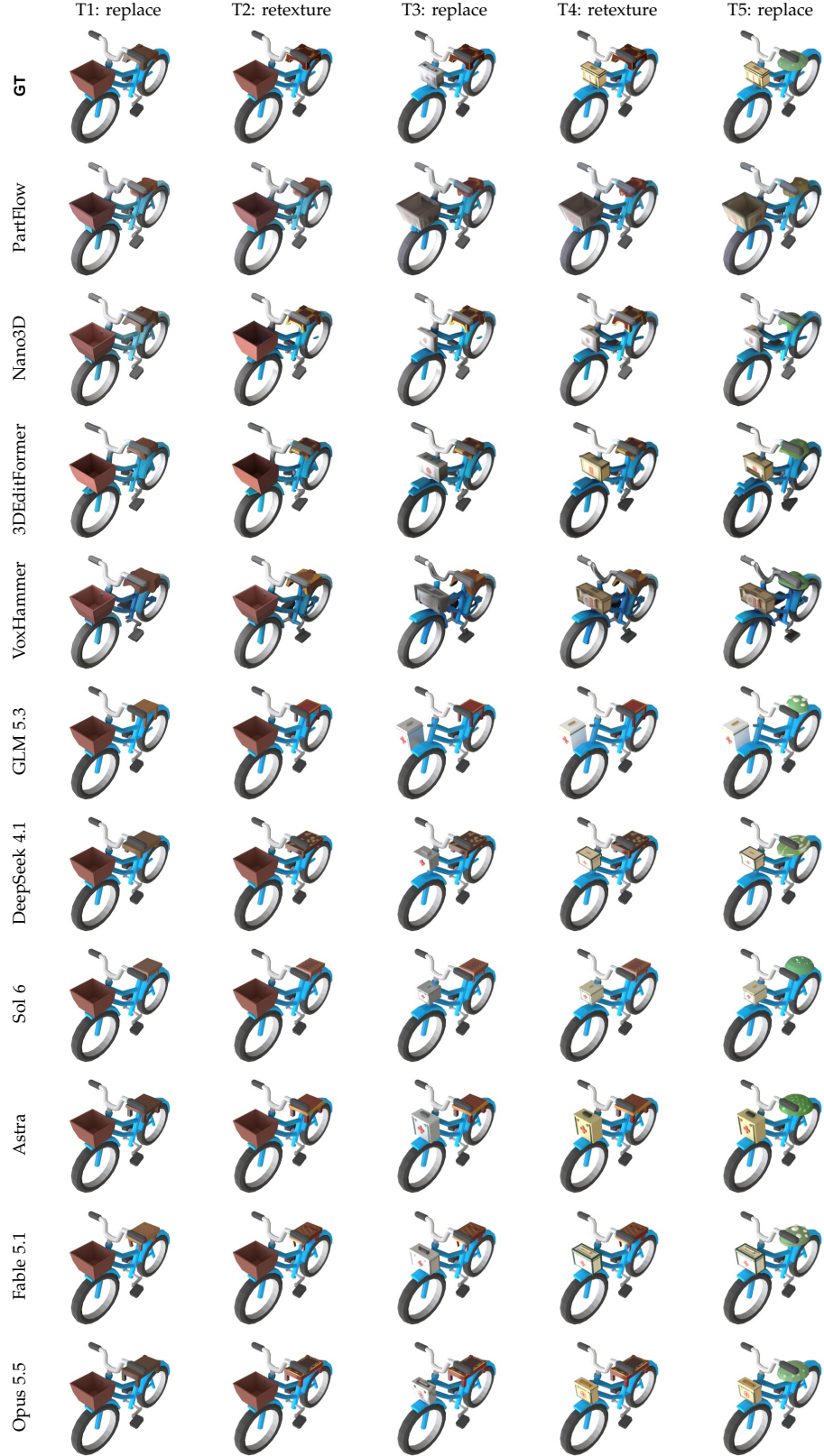}}
\caption{\textbf{A bicycle} (turns 1--5): the saddle is replaced by a wooden stool and repainted, the basket is replaced by a medical case and repainted, and the stool is replaced by a mushroom cap. Shown in the conditioning view.}
\label{fig:qual-llm-bike}
\end{figure}

\begin{figure}[p]
\centering
\ehfitpage{\input{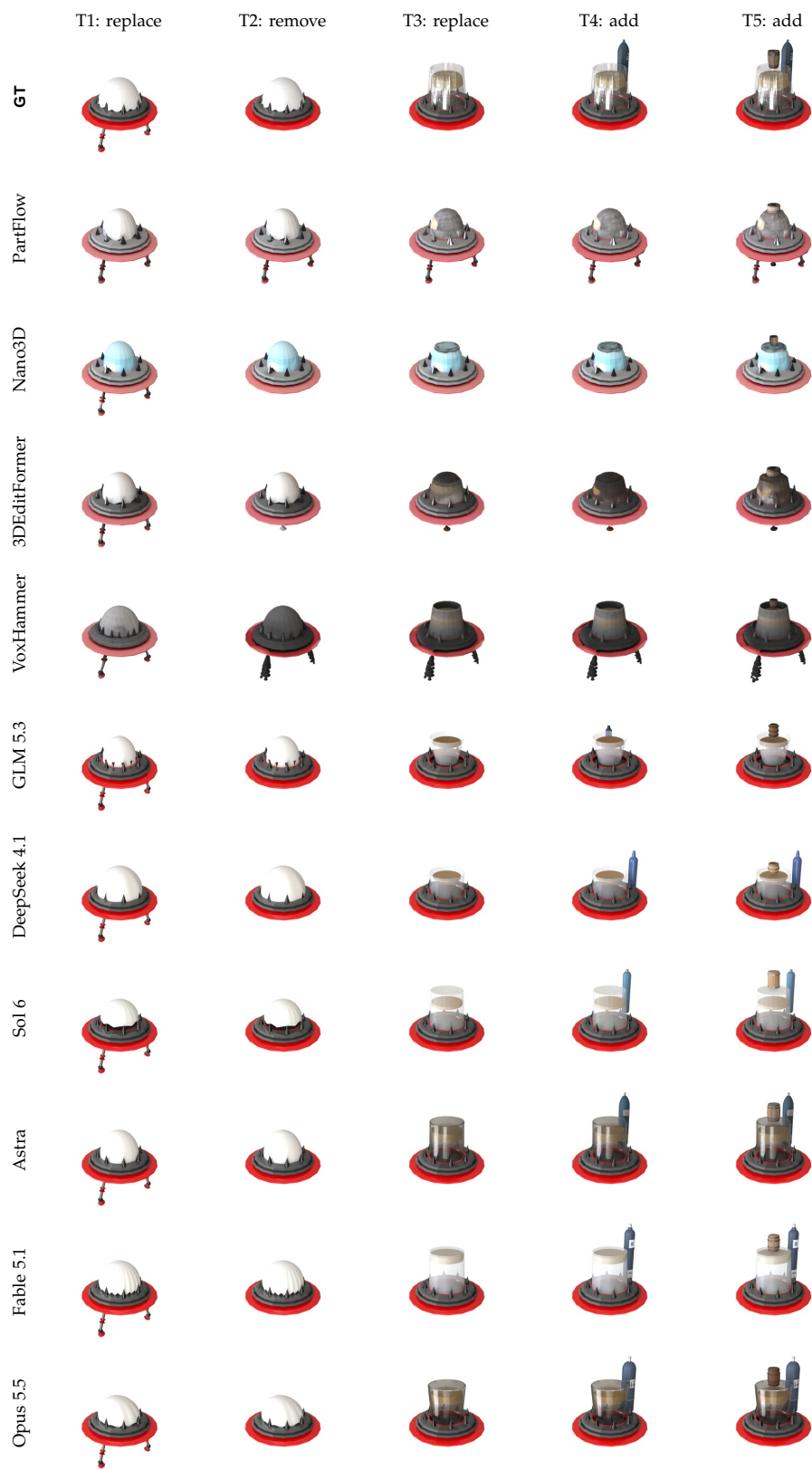}}
\caption{\textbf{A flying saucer} (turns 1--5): the dome is replaced by a parachute canopy, the landing legs are removed, the canopy is replaced by a transparent cylinder, and a gas cylinder and a barrel are added. Shown from a held-out view other than the conditioning view.}
\label{fig:qual-llm-ufo}
\end{figure}

\begin{figure}[p]
\centering
\ehfitpage{\input{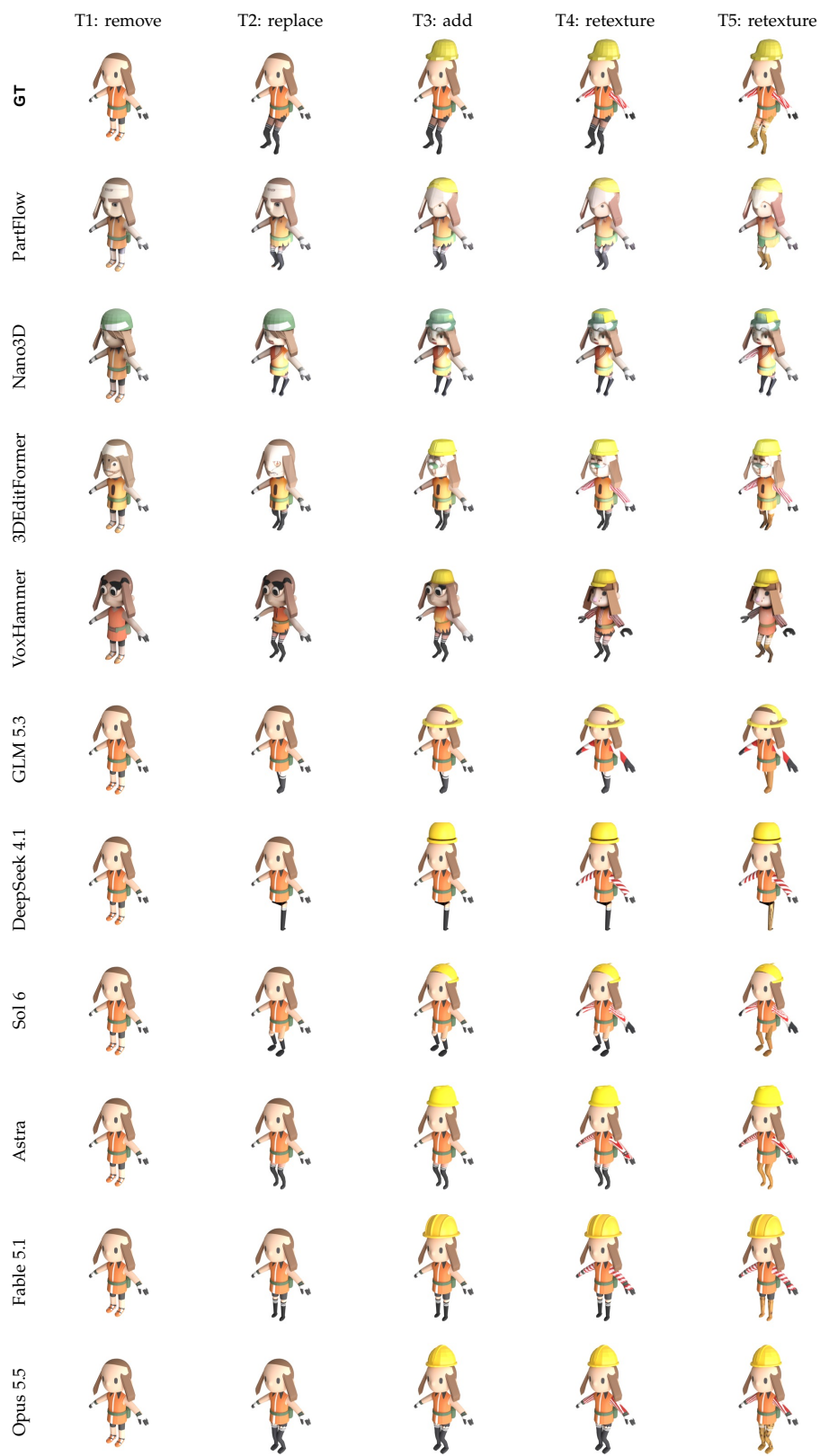}}
\caption{\textbf{A character} (turns 1--5, all 4 operations): the green cap is removed, the legs are replaced by legs in black thigh-high socks, a yellow hard hat is added, the arms are retextured with red and white candy stripes, and the legs are retextured as honey-oak wood. Shown from a held-out view other than the conditioning view.}
\label{fig:qual-character}
\end{figure}

Figures~\ref{fig:qual-llm-bike}--\ref{fig:qual-character} show 3 chains of the comparison subset with every method side by side. The 4 non-agentic methods come first, followed by the 6 agents in the order of Table~\ref{tab:agentic}. Columns are the first 5 turns and rows are methods. Every tile is the method's own output after that turn under self-rollout, rendered from the same camera as the ground truth. The non-agentic methods drift in color and shape even on turns that ask for a small change, while most agents keep their changes on the named part.

\section{Qualitative examples of chains}
\label{app:chains}
\begin{figure}[p]
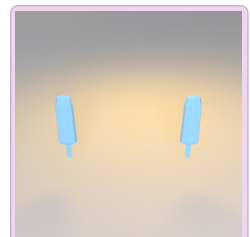

\centering
\resizebox{\linewidth}{!}{
}
\caption{Add-only and remove-only chains.}
\label{fig:app-grid-01}
\label{fig:app-style_add-only}
\label{fig:app-style_remove-only}
\label{fig:app-first}
\end{figure}

\begin{figure}[p]
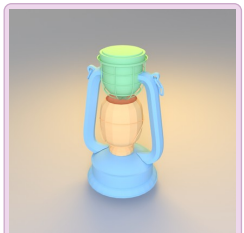

\centering
\resizebox{\linewidth}{!}{%
}
\caption{Removal and replacement chains.}
\label{fig:app-grid-02}
\label{fig:app-style_replace-only}
\end{figure}

\begin{figure}[p]
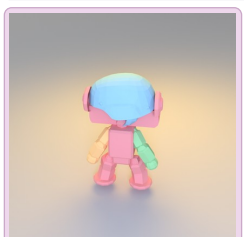

\centering
\resizebox{\linewidth}{!}{%
}
\caption{Replacement chains.}
\label{fig:app-grid-03}
\end{figure}

\begin{figure}[p]
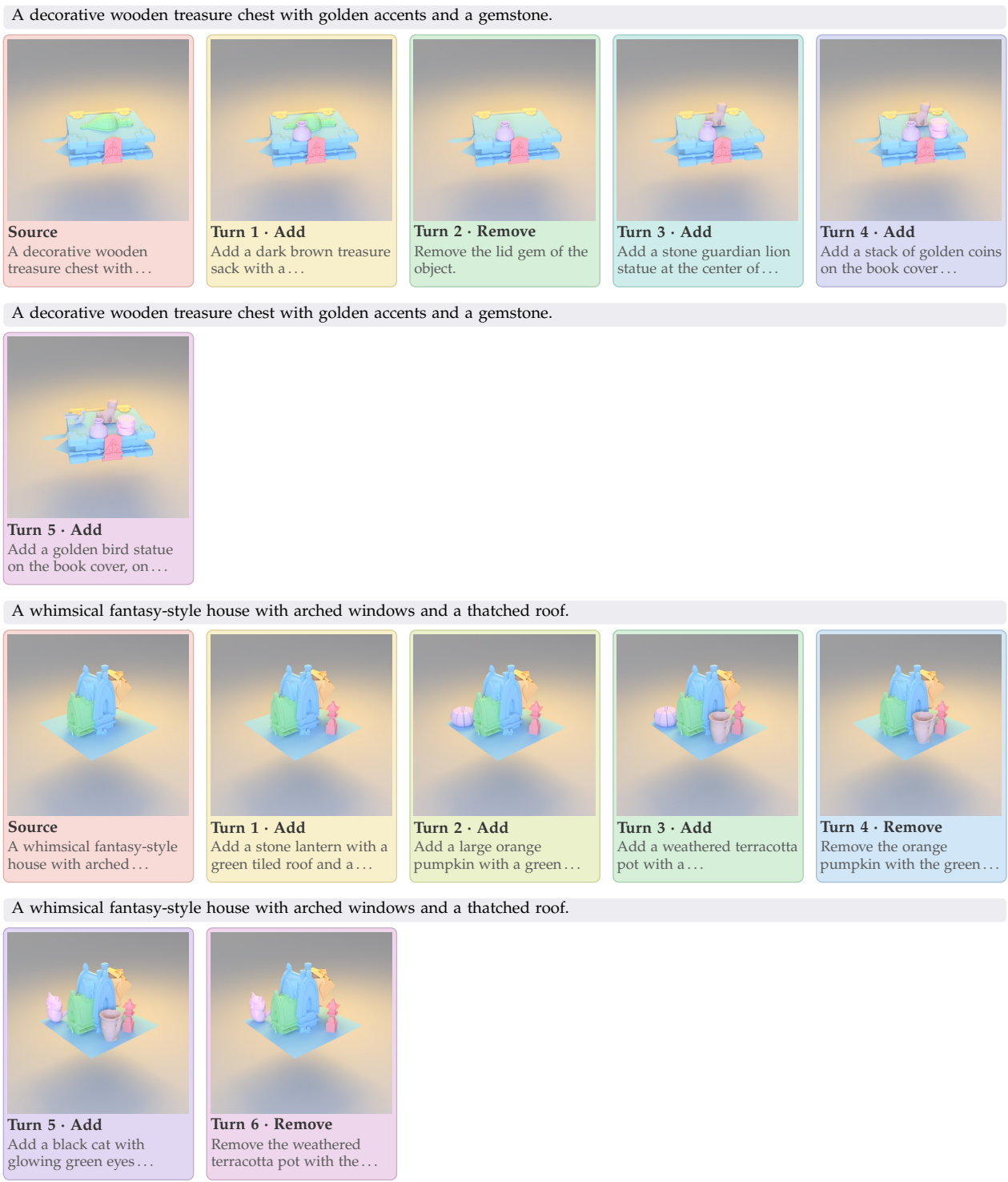

\centering
\resizebox{\linewidth}{!}{%
}
\caption{Add-and-remove chains.}
\label{fig:app-grid-04}
\label{fig:app-style_add-remove}
\end{figure}

\begin{figure}[p]
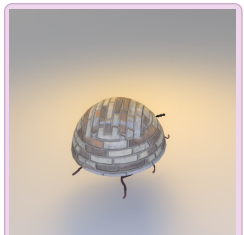

\centering
\resizebox{\linewidth}{!}{%
}
\caption{Add-and-remove and mixed-edit chains.}
\label{fig:app-grid-05}
\label{fig:app-style_mixed-geometry}
\end{figure}

\begin{figure}[p]
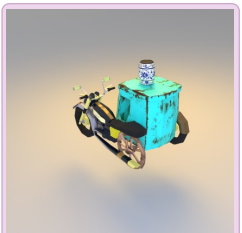

\centering
\resizebox{\linewidth}{!}{%
}
\caption{Mixed-edit chains.}
\label{fig:app-grid-06}
\label{fig:app-style_mixed-all}
\end{figure}

\begin{figure}[p]
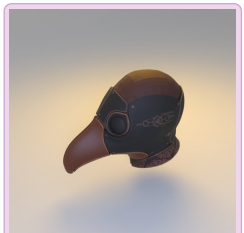

\centering
\resizebox{\linewidth}{!}{%
}
\caption{Chains with all four operations.}
\label{fig:app-grid-07}
\end{figure}

\begin{figure}[p]
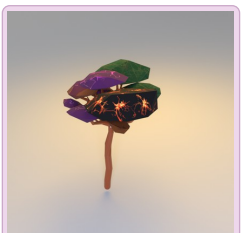

\centering
\resizebox{\linewidth}{!}{%
}
\caption{Retexture-only chains.}
\label{fig:app-grid-08}
\label{fig:app-style_retexture-only}
\end{figure}

\begin{figure}[p]
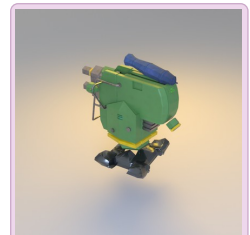

\centering
\resizebox{\linewidth}{!}{%
}
\caption{Retexture and geometry-edit examples.}
\label{fig:app-grid-09}
\label{fig:app-subject_object}
\end{figure}

\begin{figure}[p]
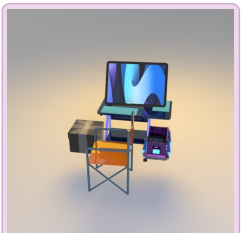

\centering
\resizebox{\linewidth}{!}{%
}
\caption{Scene editing.}
\label{fig:app-grid-10}
\label{fig:app-subject_scene}
\end{figure}

\begin{figure}[p]
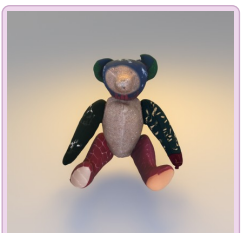

\centering
\resizebox{\linewidth}{!}{%
}
\caption{Scene and character editing.}
\label{fig:app-grid-11}
\label{fig:app-subject_humanoid}
\end{figure}

\begin{figure}[p]
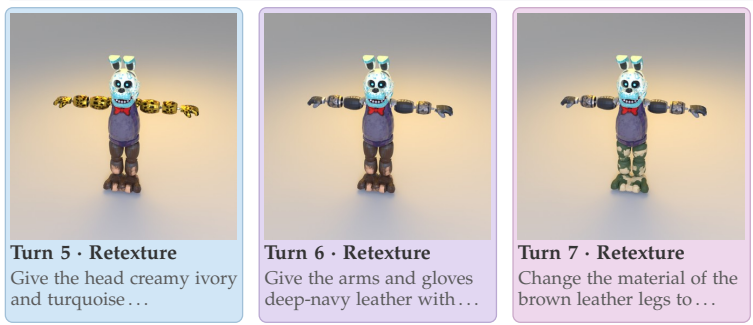

\centering
\resizebox{\linewidth}{!}{%
}
\caption{Character editing.}
\label{fig:app-grid-12}
\end{figure}

\begin{figure}[p]
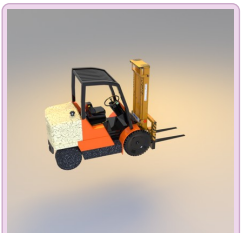

\centering
\resizebox{\linewidth}{!}{%
}
\caption{Object editing.}
\label{fig:app-grid-13}
\label{fig:app-last}
\end{figure}

Figures~\ref{fig:app-first}--\ref{fig:app-last} show ground-truth chains from EditHero. Each page uses the same grid of 5 columns and 4 rows, with equally sized cards for the source and every edited state. Each chain stays on one page and reads from left to right, continuing on the next row when needed. The operation and an instruction excerpt appear below each state. Geometry-only examples use one flat color per slot. Examples with material edits and those grouped by subject keep the assets' own textures.
\clearpage

\clearpage

\clearpage

\clearpage

\clearpage

\clearpage

\clearpage

\clearpage

\clearpage

\clearpage

\clearpage

\clearpage

\clearpage

\clearpage

\section{Prompts}\label{app:prompts}

The templates below are used for material restyling and for drafting replacement instructions, which reviewers then revised (Section~\ref{sec:benchmark}). An excerpt of the shared agent prompt is listed last.

\paragraph{Retexturing material redraw prompt}
\begin{quote}\small
Change the material to <material\_description>. Keep the shape, geometry and camera angle exactly the same. White background.
\end{quote}

\paragraph{Replacement instruction template}
\begin{quote}\small
Replace the <current\_part\_description> with <new\_part\_description>.
\end{quote}

\paragraph{LLM/VLM task prompt (excerpt).}
Each agent receives the task below, followed by the modeling skill and rendering and resource rules. These will be released with the benchmark. We quote the main task and the retexture rule, with omissions marked.
\begin{quote}\small\ttfamily\raggedright
[\dots] Edit your own previous saved output, starting with source.glb. Infer each operation and region from text/image; preserve unaffected parts. [\dots] Remove is direct physical deletion; add/replace retain the original construction and review loop; retexture directly edits only the target material/texture while preserving geometry and placement, with render-and-compare iteration. No external assets, hidden GT models/passes or generative geometry/image models. [\dots] Attempt the requested actual edit on every turn within the finite budget. Deliver the best runnable edited candidate even if quality gates fail; label it partial and continue from it. [\dots]
\end{quote}
\begin{quote}\small\ttfamily\raggedright
- Retexture: directly edit the named part's material, texture, or shader parameters to match the instruction and target appearance. Preserve its geometry, topology, UV coordinates, node names, transforms, placement, and attachment relationships. [\dots] Patterned finishes: either (a) project the target image onto the part [\dots]; or (b) generate the pattern procedurally in code [\dots] evaluated at each surface point's 3D position [\dots] rather than in UV space. Either route is acceptable; choose by render-and-compare at the evaluation camera and match the target colours.
\end{quote}

\clearpage

\end{document}